\pdfoutput=1
\documentclass[11pt]{article}

\usepackage[final]{acl}

\usepackage{times}
\usepackage{latexsym}
\usepackage[T1]{fontenc}
\usepackage[utf8]{inputenc}
\usepackage{microtype}
\usepackage{inconsolata}
\usepackage{graphicx}
\usepackage{orcidlink}

\usepackage{amsmath,amssymb}

\usepackage{booktabs}
\usepackage{array}
\usepackage{multirow}
\usepackage[dvipsnames]{xcolor}
\usepackage{tikz}
\usetikzlibrary{arrows.meta,positioning,fit}
\usepackage{listings}
\usepackage{enumitem}
\setlist[itemize]{topsep=2pt,itemsep=1pt}
\setlist[enumerate]{topsep=2pt,itemsep=1pt}
\usepackage{hyperref}

\title{fork() for Agents: Agent Kernel for \\ Non-blocking LLM Orchestration}

\author{Ankit Chadha\,\orcidlink{0000-0002-5029-0136} and Pratyay Banerjee\,\orcidlink{0000-0001-5634-410X} \\
  Amazon, AGI / Sunnyvale, USA \\
  \texttt{ankitrc, pratyay@amazon.com} \\}

\begin{document}

\maketitle

\begin{abstract}
LLM-based agents typically execute tool calls synchronously and discard completed work upon failure.
Subgraph-selective recovery, re-executing only a failed node's transitive descendants, is a well-established pattern in distributed systems (Spark, Erlang/OTP, Dryad), and recent agent runtimes (Sherlock, AgentGit, Atomix) preserve completed work through verifiers, version control, or transactional tool boundaries.
We present \textbf{Agent-Kernel}, a non-blocking runtime that ports this pattern to the LLM \emph{content plane}: its primitive, \textbf{Sub-Graph Recovery (SGR)}, fires on a raw exception with no trained verifier or planner, resets a failed node's descendants while preserving completed siblings, and composes each parent's text into its children's prompts (context inheritance), all on a durable substrate with a formal state-preservation guarantee.
Across 14{,}434 runs on synthetic and standard benchmarks (tau-bench, Gaia2, AutomationBench, InterruptBench), SGR reaches full recovery within budget=3 where whole-trajectory replay exhausts (100\% vs.\ 55\%), and is roughly 10$\times$ faster than an adapted Sherlock rollback that also recovers.
It matches ReAct-serial accuracy within noise on linear workloads while adding recovery and parallelism (up to 12.7$\times$ speedup) where structure permits. Ablation isolates SGR: removing it collapses success to 39.4\%.

\end{abstract}

\section{Introduction}\label{sec:introduction}

\begin{figure*}[t]
\centering
\includegraphics[width=0.87\textwidth]{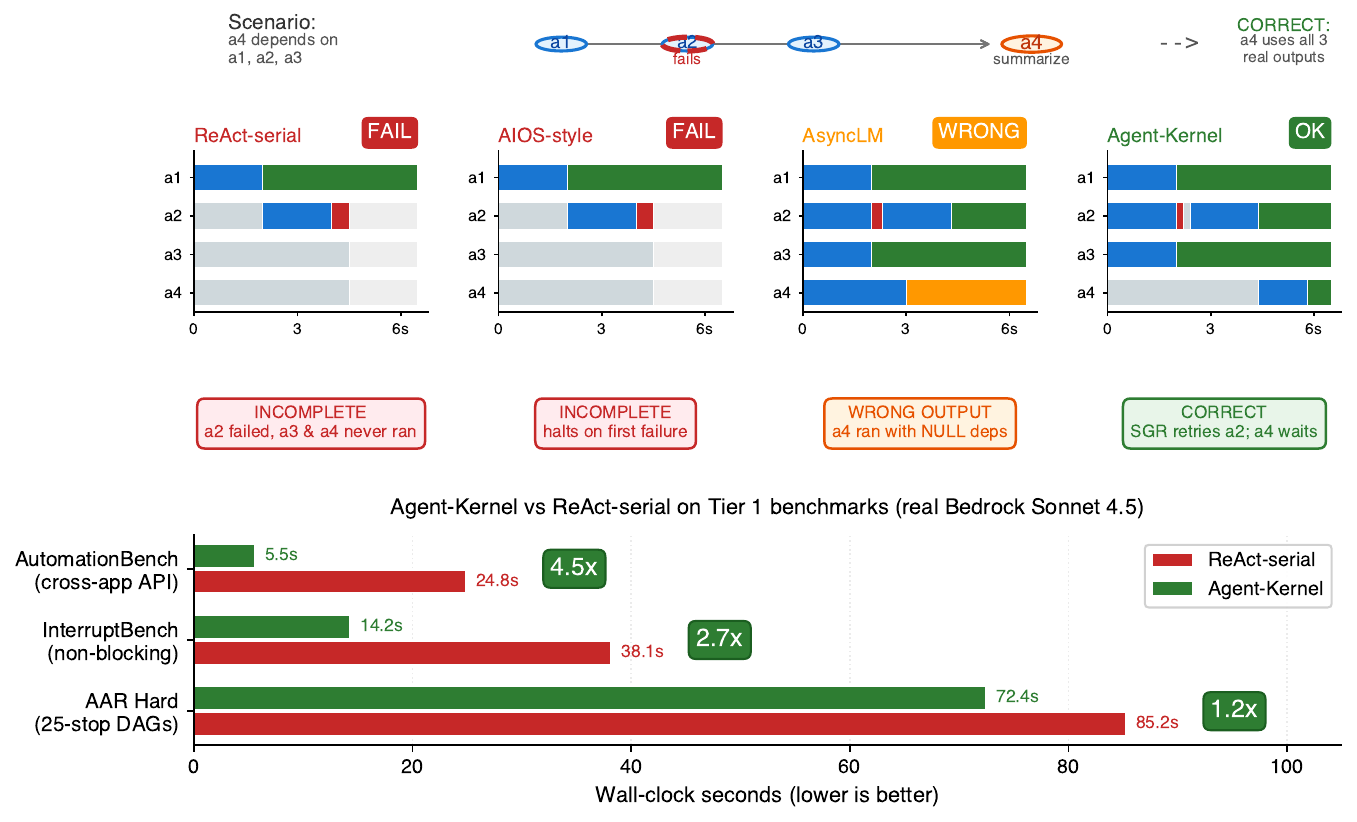}
\caption{\textbf{Top:} Dependency-sensitive scenario. Task \texttt{a4} depends on \texttt{a1,a2,a3}; \texttt{a2} fails. ReAct halts and discards all intermediate state. Agent-Kernel retries only \texttt{a2} via SGR while preserving \texttt{a1,a3}'s completed outputs, then dispatches \texttt{a4} with correct parent context. \textbf{Bottom:} Wall-clock speedup on real-inference benchmarks (Sonnet 4.5): 4.5$\times$ on AutomationBench (cross-app API orchestration), 2.7$\times$ on InterruptBench (non-blocking main loop). Speedups reflect latency reduction, not accuracy improvement.}
\label{fig:teaser}
\end{figure*}

LLM-based autonomous agents excel at multi-step reasoning and tool use~\citep{yao2023react,wu2023autogen}, yet their runtime architectures remain fundamentally synchronous: each tool call blocks the agent's main loop until completion, and any failure triggers either a full halt or a costly whole-trajectory replay that discards all intermediate progress. As agents move from research prototypes to production workflows orchestrating dozens of sub-agents across real-world services, system-level reliability primitives become a first-class research need~\citep{trivedi2024appworld}. The current synchronous, all-or-nothing execution model is the computational equivalent of a single-process operating system before \texttt{fork}: one task at a time, no concurrency, no selective recovery. (We use OS terminology metaphorically: \texttt{fork} creates a DAG task node, not a process clone; ``kernel'' denotes the scheduling substrate.)

Recent work addresses this from multiple angles. AsyncLM~\citep{gim2024asynclm} introduces token-level interrupts for non-blocking function calls but ignores declared dependencies between concurrent calls. AIOS~\citep{mei2024aios} provides OS-inspired scheduling at the agent level but lacks reactive cascading or failure recovery. DynTaskMAS~\citep{dyntaskmas2025} constructs dynamic task graphs but relies on a planner LLM to emit the DAG, making dependency correctness contingent on planner accuracy, and replans the entire remaining graph on failure rather than retrying the minimal affected subgraph. Three recent runtimes preserve completed work under failure: AgentGit~\citep{li2025agentgit} adds Git-style commit, revert, and branching to LangGraph; Atomix~\citep{mohammadi2026atomix} adds transactional tool-call boundaries with commit/abort and compensation; and Sherlock~\citep{ro2025sherlock}, the closest analog, rolls a task DAG back to its last verified state on verifier disagreement, using a trained analyzer, a verifier selector, and an LLM planner.

\textbf{Selective subgraph recovery is not new; the content plane is.} Resetting a failed node's transitive descendants while preserving completed siblings is a well-established pattern in distributed systems: Spark's RDD lineage recomputes lost partitions~\citep{zaharia2012rdd}, Erlang/OTP's \texttt{rest\_for\_one} restarts a failed process and those started after it~\citep{armstrong2003erlang}, and Dryad re-runs failed vertices and their dependents~\citep{isard2007dryad}. We do not claim this pattern as our contribution. What these systems (and Atomix, AgentGit, and Sherlock) reason about is the \emph{control and state} of execution: dependencies and side effects, versions, verified outputs. LLM orchestration adds a \emph{content plane} they do not expose as a first-class mechanism: an upstream node's variable-length text is composed into a downstream prompt under a token budget (context inheritance), and a retry may rewrite that prompt on format drift rather than blindly re-executing. Our contribution is that plane, delivered as one runtime with a formal state-preservation guarantee and measured recovery speed. Concretely: no prior LLM orchestration system provides exception-triggered, zero-shot, DAG-native sibling preservation without installing a verifier or planner, and Agent-Kernel recovers roughly 10$\times$ faster than \emph{our} adapted Sherlock rollback on our synthetic DAG grids (parametrized by parallel width $K$ and depth $L$; here $K{=}8$, $L{=}8$). This gap reflects that adaptation's serial replay and per-output verifier pass, not necessarily an optimally-engineered Sherlock (\S\ref{sec:related-work}, App.~\ref{app:sherlock}).

We present \textbf{Agent-Kernel}, a non-blocking orchestration runtime for LLM agents that realizes this content plane. Our contributions are threefold:

\begin{enumerate}[leftmargin=*,itemsep=2pt]
\item \textbf{Sub-Graph Recovery (SGR)}: an exception-triggered, zero-shot retry primitive that resets a failed node's transitive-closure subgraph while preserving completed siblings under a formal state-preservation guarantee, with LLM content-plane semantics (context inheritance, and a prompt-mutation retry extension).
\item A \textbf{durable runtime substrate}: a SQLite-backed Sub-Agent Buffer (SAB) with a 7-state FSM and a reactive DAG State Propagator (DSP) that makes SGR safe across process boundaries.
\item An \textbf{empirical study} (\S\ref{sec:experiments}) showing SGR recovers where whole-trajectory replay exhausts, is faster than baselines that also recover fully, matches ReAct-serial accuracy on linear workloads, and, unlike shipped LangGraph/AutoGen/LlamaIndex, recovers without user configuration.
\end{enumerate}

\textbf{Scope.} Agent-Kernel is not a planner (the user declares dependencies via a 5-verb API), not a policy trainer (we use a frozen Sonnet 4.5 backbone), not a failure detector (SGR fires on whatever signal marks a node failed; in our experiments, a raised exception), and does not claim state-of-the-art task accuracy. Our contribution is the orchestration substrate: it composes with any LLM backend, and its gains are orthogonal to policy-level improvements from fine-tuning or RL. As agents scale to multi-step workflows over dozens of sub-agents, the field needs system-level primitives for concurrency and recovery, much as operating systems provide for processes. Agent-Kernel adapts one such established pattern, giving it LLM content-plane semantics as one measured runtime with a formal guarantee.

\section{Related Work}\label{sec:related-work}

Subgraph-selective recovery is an established pattern, and recent LLM runtimes preserve completed work under failure. We therefore position Agent-Kernel by \emph{what plane it operates on} rather than by novelty of the recovery pattern. Table~\ref{tab:related} contrasts the three closest agent runtimes with Agent-Kernel along recovery trigger, recovery action, and whether the system composes upstream text into downstream prompts (context inheritance).

\begin{table*}[t]
\centering
\small
\resizebox{\textwidth}{!}{%
\begin{tabular}{llll c}
\toprule
System & Recovery trigger & Recovery action & Config-free & Content-plane policy \\
\midrule
Atomix~\citep{mohammadi2026atomix} & Fault or conflict & Retry/wait/abort + compensation; frontiers isolate work & Yes & None \\
AgentGit~\citep{li2025agentgit} & User- or agent-invoked & Revert or branch to a saved state (Git-style) & No & None \\
Sherlock~\citep{ro2025sherlock} & Verifier disagreement & Roll back to last verified output & No (verifier+planner) & None \\
\textbf{Agent-Kernel (ours)} & \textbf{Exception (zero-shot)} & \textbf{Reset descendants, keep siblings, re-execute} & \textbf{Yes} & \textbf{Budgeted + mutable} \\
\bottomrule
\end{tabular}%
}
\caption{Agent-Kernel against the three closest LLM agent runtimes. All three preserve completed work. Agent-Kernel's distinguishing properties are an exception-triggered, config-free trigger requiring no verifier or planner, and an explicit content-plane \emph{policy}: it composes upstream text into downstream prompts under a token budget and may rewrite a prompt on format drift---operations these systems, which reason over control and state, do not expose as a first-class mechanism.}
\label{tab:related}
\end{table*}

\textbf{The three closest analogs.} \textbf{Sherlock}~\citep{ro2025sherlock} is the nearest: it rolls a task DAG back to its last verified state, resetting the failed node's descendants, but it fires on \emph{verifier disagreement} and requires a trained analyzer, a verifier selector, and an LLM planner, whereas SGR fires on a raw exception with no such setup. We evaluate a paper-faithful \emph{Sherlock-lite} (cannot fire on an exception) and a steelmanned \emph{Sherlock-generous} (treats an exception as verifier disagreement); the latter recovers fully, so SGR is not the only mechanism that recovers under our failure model, but SGR is roughly 10$\times$ faster because it re-dispatches only children whose parents just completed rather than re-executing the transitive closure serially and re-running the verifier on every intermediate output (App.~\ref{app:sherlock}). \textbf{Atomix}~\citep{mohammadi2026atomix} operates one layer below us, at the tool-call boundary: it tags calls with epochs, tracks per-resource frontiers, and commits only when no conflicting work can arrive, compensating external effects on abort. Both systems use declared dependencies, an advancing frontier, and recovery instead of halting; the difference is the plane: Atomix's unit is the tool call, aimed at external-effect consistency, and it scopes out context composition, so the two layers compose. \textbf{AgentGit}~\citep{li2025agentgit} provides Git-style commit, revert, and branching on LangGraph with reuse across trajectories; it is user- or agent-invoked and, being built on LangGraph, does not provide automatic transitive-closure reset with sibling preservation, nor context inheritance or prompt-mutation retry.

\textbf{Prompt-mutation retry vs.\ self-correction.} Our prompt-mutation retry, re-issuing a failed node's prompt augmented with the error and expected schema, is a form of error-feedback self-correction, and we claim no novelty for that idea: Reflexion~\citep{shinn2023reflexion} and Self-Refine~\citep{madaan2023selfrefine} established verbal self-feedback loops for single agents. What is new is placing this operation \emph{inside} the recovery cascade of a DAG runtime: the mutated retry is applied only to the reset subgraph while completed siblings' outputs are preserved and inherited, so self-correction and selective recovery compose rather than being separate mechanisms.

\textbf{SGR's systems lineage.} Resetting a failed node's transitive descendants while preserving completed siblings is well established: Spark's RDD lineage recomputes lost partitions by tracing the dependency graph~\citep{zaharia2012rdd}; Erlang/OTP's \texttt{rest\_for\_one} restarts a crashed process and those started after it~\citep{armstrong2003erlang}; Dryad re-runs failed vertices and their downstream dependents~\citep{isard2007dryad}; and Dichev et al.~\citep{dichev2017rollback} formalize dependency-aware rollback for HPC task runtimes. The Dual-State Architecture~\citep{dualstate2025} proposes cascade invalidation for LLM agents but triggered by downstream validation failure, resetting \emph{upstream} steps. We claim no novelty for the pattern itself; our adaptation is to the LLM content plane, where node outputs are variable-length text injected into downstream prompts and a retry may rewrite that prompt on format drift.

\textbf{Other orchestration approaches.} ReAct~\citep{yao2023react} halts on failure. AsyncLM~\citep{gim2024asynclm} retries individual calls but ignores dependency ordering. DynTaskMAS~\citep{dyntaskmas2025} replans the entire remaining graph on failure. Production frameworks such as LangGraph~\citep{chase2024langgraph}, AutoGen~\citep{wu2023autogen}, and LlamaIndex~\citep{liu2022llamaindex} support graph-like coordination but, by default, halt or require full state rewind on a node failure; LangGraph's default retry filter also excludes the application-level exceptions we inject (\S\ref{sec:rq3}, App.~\ref{app:integration}). Workflow engines (Airflow~\citep{airflow2015}, Temporal~\citep{temporal2020}, Prefect~\citep{prefect2018}, Parsl~\citep{parsl2019}) persist DAGs and support sub-DAG retry, but assume fixed schemas and deterministic tasks and do not model token budgets, prompt construction, or stochastic format drift. Native parallel tool-use~\citep{kim2024llmcompiler} covers width-1 parallelism within a turn but not multi-level DAGs, cross-turn recovery, or interrupts.

\textbf{Benchmarks motivating non-blocking agents.} tau-bench~\citep{yao2024taubench} evaluates multi-turn tool-agent-user interaction under serial execution. Gaia2~\citep{froger2026gaia2} introduces asynchronous environments where world events occur independently of the agent, motivating main-loop responsiveness. InterruptBench~\citep{interruptbench2026} evaluates requests arriving mid-execution, the scenario Agent-Kernel's non-blocking \texttt{fork} targets, but provides no recovery mechanism. The Amazing Agent Race~\citep{aar2026} stress-tests multi-tool orchestration on trail-navigation tasks with naturally parallel API calls. These benchmarks establish concurrency and failure resilience as first-class requirements for production LLM agents.

\section{Method}\label{sec:method}

Agent-Kernel comprises a durable runtime substrate (SAB + DSP) and a recovery primitive (SGR); Figure~\ref{fig:architecture} (App.~\ref{app:architecture}) shows the dataflow. We present SGR first, then the substrate enabling it.

\subsection{Sub-Graph Recovery (SGR)}\label{sec:sgr}

When a node fails, the failure cascade first marks its transitive descendants \texttt{unreachable}. A budget-bounded control loop (enforced by the caller via the node's attempt history, App.~\ref{app:artifacts}) then invokes \texttt{retry} on the failed node while budget remains; SGR resets that node and its unreachable descendants to \texttt{waiting} and re-runs them:

\begin{lstlisting}[basicstyle=\ttfamily\scriptsize,breaklines=true]
subgraph_retry(node_id):
    assert state(node_id) == FAILED
    for d in transitive_closure(node_id):
        if state(d) == UNREACHABLE:      # = every non-complete descendant in F
            transition(d, WAITING)
    re_evaluate(node_id)  # -> WAITING/RUNNING; forward cascade re-runs the rest
\end{lstlisting}

\textbf{State-preservation guarantee.} Let $G = (V, E)$ be the task DAG, $F \subset V$ the failed subgraph (failed node + transitive descendants), and $C = V \setminus F$ the complement. SGR guarantees: $\forall c \in C$, \texttt{state(c)} and \texttt{outputs(c)} are unchanged after retry---completed siblings are never re-executed. No node in $F$ can be \texttt{complete} when SGR fires (a strict descendant completes only after its ancestor, contradicting the ancestor being \texttt{failed}), so its non-complete descendants are exactly the \texttt{unreachable} ones marked by the failure cascade, and resetting them discards no completed work. Re-execution is level-triggered: DSP re-dispatches any reset node whose parents (all in $C$) are \texttt{complete}, and the forward-success cascade carries the rest. The retry budget is enforced by the caller (the failed node is retried only while its recorded error count is below budget), not inside SGR.

\begin{figure}[t]
\centering
\begin{tikzpicture}[
  font=\sffamily\footnotesize,
  nd/.style={circle,draw,line width=0.9pt,minimum size=5.2mm,inner sep=0pt},
  rs/.style={nd,draw=violet!80!black,fill=violet!12},
  kp/.style={nd,draw=green!45!black,fill=green!45!black,text=white},
  rx/.style={nd,draw=red!70!black,fill=red!10},
  lab/.style={font=\sffamily\scriptsize,inner sep=1pt},
  ar/.style={-{Stealth[length=3.5pt]},black!65,line width=0.6pt}]
\node[rs] (a1) at (0,0) {a};
\node[kp] (b1) at (1.0,0) {b};
\node[kp] (c1) at (2.0,0) {c};
\node[rs] (s1) at (1.0,-1.15) {s};
\foreach \u in {a1,b1,c1} \draw[ar] (\u) -- (s1);
\node[lab,violet!80!black,above=1pt of a1] {failed};
\node[lab,green!35!black,above=1pt of b1] {kept};
\node[lab,green!35!black,above=1pt of c1] {kept};
\node[font=\sffamily\small\bfseries,green!35!black] at (1.0,0.85) {SGR};
\node[lab,align=center,below=3pt of s1] {re-executes \textbf{a, s}\\keeps \textbf{b, c} (outputs in SAB)};
\draw[black!25] (2.75,1.0) -- (2.75,-2.05);
\node[rx] (a2) at (3.5,0) {a};
\node[rx] (b2) at (4.5,0) {b};
\node[rx] (c2) at (5.5,0) {c};
\node[rx] (s2) at (4.5,-1.15) {s};
\foreach \u in {a2,b2,c2} \draw[ar] (\u) -- (s2);
\node[lab,red!70!black,above=1pt of a2] {failed};
\node[lab,red!70!black,above=1pt of b2] {re-run};
\node[lab,red!70!black,above=1pt of c2] {re-run};
\node[font=\sffamily\small\bfseries,red!70!black] at (4.5,0.85) {FullReplan};
\node[lab,align=center,below=3pt of s2] {re-executes \textbf{a, b, c, s}\\keeps nothing};
\end{tikzpicture}
\caption{After \texttt{a} fails in the worked example (\S\ref{sec:api}), SGR resets only \texttt{a} and its descendant \texttt{s}; completed siblings \texttt{b}, \texttt{c} keep their outputs and are inherited by \texttt{s}. Whole-task replay re-runs every node, and each re-run is another chance for a transient failure to spend budget. On the failure grid this is the gap between SGR (100\%) and matched-budget FullReplan (56.2\%, Table~\ref{tab:ablation-compact}).}
\label{fig:sgr-vs-replay}
\end{figure}
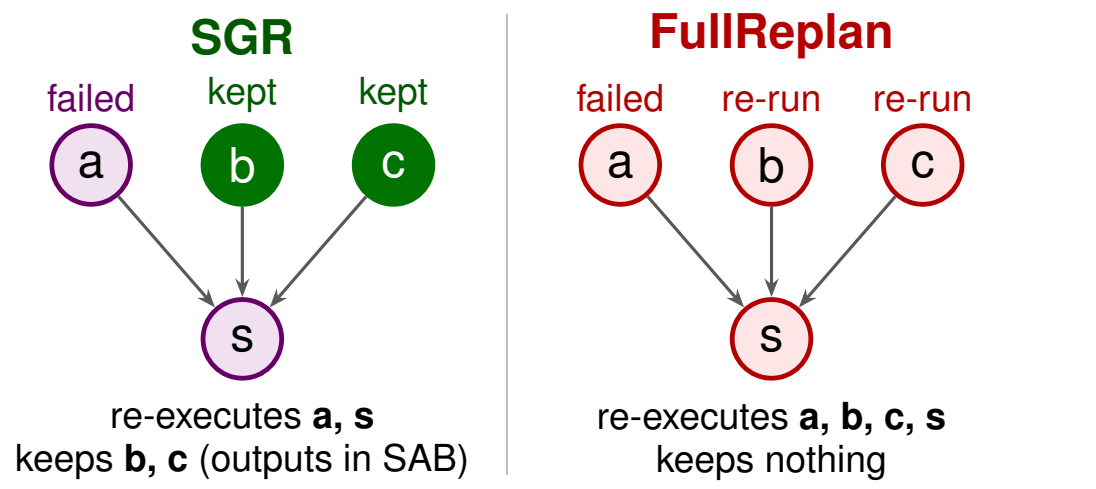

\textbf{Why this beats full replay.} Whole-trajectory replay (same budget $k$) restarts from the root, re-executing all $d$ completed layers before reaching a depth-$d$ failure; a new transient failure in any of them decrements the budget without progress. SGR skips completed layers, spending budget only on the failure point; hence 100\% vs FullReplan's 56.2\% under matched budgets (Figure~\ref{fig:sgr-vs-replay}, Section~\ref{sec:rq3}).

\textbf{Detection is separate from recovery.} SGR consumes a failure signal; it does not decide what counts as a failure. In our experiments the signal is a raised exception, the controlled trigger we use to isolate retry scope. Any detector that marks a node \texttt{failed} (a schema or parameter validator, a tool-call checker, or a verifier's disagreement as in Sherlock) fires the same cascade and the same reset, so SGR can run underneath a verifier unchanged. The prompt-mutation probe (App.~\ref{app:pmr-ablation}) is one such composition: a schema check detects format drift, and the mutated retry runs inside the reset subgraph.

\textbf{Scope of the guarantee.} The guarantee covers the states and outputs that SAB records, not effects a node has already had on the outside world: if a non-idempotent tool call (a payment, a file write, a sent message) succeeds and its node then fails, re-executing the node can repeat the effect. Guarding such effects is the job of a transactional tool boundary one layer below SGR, such as Atomix~\citep{mohammadi2026atomix}, which tags calls and compensates external effects on abort; the two layers compose, with SGR resetting content-plane state and the tool boundary guarding external effects. Within SAB, every state transition is a single SQL transaction, so concurrent retries observe a consistent state, and a reset descendant that also depends on another still-failed node stays \texttt{waiting} until that parent completes, since dispatch requires all parents to be \texttt{complete}. The crash test (App.~\ref{app:crash}) checks durability: completed nodes survive a process kill and stale \texttt{running} nodes are re-queued.

\subsection{Runtime Substrate: SAB + DSP}\label{sec:substrate}

SGR requires two properties from its substrate: (1) durable storage of node states and outputs (so completed work survives crashes and retries), and (2) reactive scheduling (so reset nodes are automatically re-dispatched when their parents are complete).

\textbf{Sub-Agent Buffer (SAB).} A SQLite database (WAL mode) tracking each sub-agent through a 7-state FSM: \texttt{pending} $\rightarrow$ \texttt{waiting} $\rightarrow$ \texttt{running} $\rightarrow$ \texttt{complete}|\texttt{failed}|\texttt{cancelled}|\texttt{unreachable}. State transitions are atomic (single SQL transaction). An \texttt{edges} table stores dependency pairs for O(1) traversal during SGR's transitive-closure computation. A \texttt{failed} node returns to \texttt{waiting} via \texttt{retry}; the budget bound (whether to retry at all) is applied by the control loop, not the FSM. Each record stores:

\noindent\begin{minipage}{\columnwidth}
\begin{lstlisting}[basicstyle=\ttfamily\scriptsize,breaklines=true]
SubAgentRecord {
    sub_agent_id:  str        -- primary key
    task:          str        -- the prompt/goal
    role:          str        -- prompt persona
    state:         State      -- 7-state FSM
    depends_on:    list[str]  -- parent sub-agent IDs
    intermediate_outputs: list[dict]  -- partial/final results
    attempt_history: list[dict]       -- start/error events (errors drive retry)
    retry_budget:  int        -- max attempts (benchmarks: 3)
}
\end{lstlisting}
\end{minipage}

\textbf{DAG State Propagator (DSP).} A background daemon watching SAB transitions and propagating four cascade types: (1)~\emph{Forward-success}: when a node completes, evaluate its dependents' readiness; dispatch ready nodes with parents' outputs injected as prompt context. (2)~\emph{Failure}: when a node fails, mark its transitive descendants \texttt{unreachable} immediately. (3)~\emph{Retry}: reset failed subgraph (SGR's entry point), invoked by a budget-bounded control loop. (4)~\emph{Cancel}: mark descendants \texttt{unreachable}.

\textbf{Context inheritance and prompt-mutation retry (the content plane).} When DSP dispatches a child, it reads each declared parent's \texttt{intermediate\_outputs} from SAB and composes them into the child's prompt as a structured block under a token budget. This ensures children see their parents' actual outputs, not stale or missing data: the correctness property that unordered parallelism violates. As a design extension (evaluated as a preliminary probe, not part of the shipped recovery loop), a retry can additionally rewrite a node's prompt with error context (the previous malformed output and expected schema) rather than re-issuing it unchanged. These two operations (composing variable-length upstream text under a budget, and prompt-mutation on drift) are the LLM-specific plane that control-and-state runtimes (Atomix, AgentGit, Sherlock) do not expose as a first-class policy; App.~\ref{app:contentplane} ablates each.

\subsection{The 5-Verb API}\label{sec:api}

Users interact through five verbs.

\begin{table}[t]
\centering
\small
\setlength{\tabcolsep}{5pt}
\resizebox{\columnwidth}{!}{%
\begin{tabular}{@{}lll@{}}
\toprule
Verb & Effect & Cascade type \\
\midrule
\texttt{fork} & Creates sub-agent, returns immediately & spawn \\
\texttt{retry} & Resets \& re-dispatches failed subgraph & SGR reset \\
\texttt{cancel} & Marks descendants unreachable & cancel cascade \\
\texttt{query} & Returns matching SAB records & set-oriented read \\
\texttt{fetch} & Reads one record's current outputs & per-record read \\
\bottomrule
\end{tabular}%
}
\caption{The 5-verb API and the scheduler cascade each verb drives (\texttt{fork} is the \texttt{spawn} primitive; full call signatures in App.~\ref{app:verb-signatures}). The main loop returns immediately after each \texttt{fork}.}
\label{tab:api}
\end{table}

Each verb maps to a distinct scheduler cascade (Table~\ref{tab:api}); \texttt{retry} and \texttt{fork} are not redundant: \texttt{fork} creates new descendants, whereas \texttt{retry} resets and re-dispatches existing ones. The main loop never blocks on sub-agent execution. DSP handles all scheduling via the forward-success cascade: when a node's parents all reach \texttt{complete}, DSP automatically transitions it to \texttt{running} and dispatches execution with parents' outputs injected. The synthetic harness exercises only \texttt{fork} and \texttt{retry}; App.~\ref{app:sre} gives an SRE incident-response scenario that exercises all five verbs at 100\% invocation.

\textbf{Worked example.} A user wants to compare prices from three vendors, then summarize:

\begin{lstlisting}[basicstyle=\ttfamily\scriptsize,breaklines=true]
a = fork("Get price from Vendor A", role="researcher")
b = fork("Get price from Vendor B", role="researcher")
c = fork("Get price from Vendor C", role="researcher")
s = fork("Compare and recommend", role="analyst",
          depends_on=[a, b, c])
# Main loop returns immediately. If a fails:
retry(a)  # SGR resets a and s, preserves b and c
\end{lstlisting}

DSP transitions \texttt{s} to \texttt{running} when \texttt{a}, \texttt{b}, and \texttt{c} all complete. If \texttt{a} fails, it stays \texttt{failed} and the failure cascade immediately marks \texttt{s} \texttt{unreachable}. The control loop, seeing \texttt{a} failed with budget remaining, calls \texttt{retry(a)}: SGR resets \texttt{a} to \texttt{waiting} and its \texttt{unreachable} descendant \texttt{s} back to \texttt{waiting}, while preserving \texttt{b} and \texttt{c}'s completed outputs; the forward cascade re-runs \texttt{a} then \texttt{s}.

\section{Experiments}\label{sec:experiments}

\subsection{Setup}\label{sec:setup}

\textbf{Model and infrastructure.} Primary experiments use Claude Sonnet 4.5 via LLM API, round-robin across 3 regions; cross-backbone checks add Haiku 4.5 and Opus 4.5. Total evaluation: 14,434 runs (11,809 in the core grids and benchmarks, 2,625 in the comparator and scaling studies); a per-evaluation footprint is in App.~\ref{app:footprint} (Table~\ref{tab:eval-scale-compact}).

\textbf{Systems.} All systems share the same executor, model, prompt format, and failure-injection wrapper:
(1)~\textbf{ReAct-serial}~\citep{yao2023react}: single-thread, halts on failure;
(2)~\textbf{ReAct+retry}: whole-trajectory replay, budget=3;
(3)~\textbf{ReAct-per-step-retry (B3)}: persists failed observations in context and retries the individual step, a stronger baseline that recovers fully but runs serially;
(4)~\textbf{AsyncLM-style}~\citep{gim2024asynclm}: fan-out ignoring dependencies, retry-once;
(5)~\textbf{AIOS-style}~\citep{mei2024aios}: round-robin tick scheduler;
(6)~\textbf{Sherlock-lite / Sherlock-generous}~\citep{ro2025sherlock}: paper-faithful and steelmanned adaptations of DAG rollback-to-verified-state (App.~\ref{app:sherlock});
(7)~\textbf{Agent-Kernel}: full SAB+DSP+SGR.
Ablation variants: NoSAB, NoDSP, NoSGR, plus FullReplan (whole-task restart, same budget; the ablation analog of ReAct+retry, both replaying from the root).

\textbf{Framework comparison methodology.} We report two complementary comparisons. (i)~\emph{Primitive re-implementations} (§\ref{sec:rq3}, App.~\ref{app:frameworks}) reproduce each framework's default \emph{scheduling strategy} (full-rewind, halt, conversation-driven retry) on our synthetic grid; they test the strategy, not the binary. (ii)~\emph{Real-binary integration tests} (App.~\ref{app:integration}) run installed LangGraph~1.2, LlamaIndex~0.14, and AutoGen~v0.4 on a diamond fan-out DAG with one failing leaf. We label the two clearly throughout and never present (i) as a benchmark of the production systems.

\textbf{Benchmarks.} \emph{Synthetic}: 120 configurations (parallel width K$\in$\{1,2,4,8,16\} $\times$ depth L$\in$\{1,2,4,8\} $\times$ failure\_rate$\in$\{0,0.1,0.3\} $\times$ interrupt on/off) $\times$ 3 seeds = 360 runs/system; 1,440 across the four scheduling modes. \emph{External (real LLM inference)}: AutomationBench~\citep{automationbench2026}, AAR~\citep{aar2026}, InterruptBench~\citep{interruptbench2026}, AppWorld~\citep{trivedi2024appworld} (wall-clock in Table~\ref{tab:external-wallclock}). \emph{Linear baselines}: tau-bench~\citep{yao2024taubench} (1,320 runs), Gaia2~\citep{froger2026gaia2} (2,560 runs). The comparator and scaling studies (Sherlock/B3 grids, cross-backbone, real-framework integration, K-sweep) are detailed in the appendix.

\subsection{Selective Retry vs.\ Full Replay}\label{sec:rq1}

\begin{table}[t]
\centering
\small
\setlength{\tabcolsep}{2.5pt}
\resizebox{\columnwidth}{!}{%
\begin{tabular}{@{}l r@{\,}l r l@{}}
\toprule
System & \multicolumn{2}{c}{Recovery} & Wall & Strategy \\
\midrule
\textbf{Agent-Kernel} & \textbf{100.0\%} & \textbf{[97.0, 100.0]} & \textbf{21s} & SGR \\
ReAct-per-step (B3)   & 100.0\% &                & 135s & Per-step \\
Sherlock-gen.$^{\S}$    & 100.0\% &          & 149s & Rollback \\
AsyncLM$^\dagger$     & 100.0\% & [97.0, 100.0]  & --- & Retry-once \\
ReAct+retry                 & 55.0\%  & [46.1, 63.6]   & 15s & Full replay \\
ReAct-serial                & 36.7\%  & [28.5, 45.7]   & 3s & Halt \\
AIOS-style                  & 36.7\%  & [28.5, 45.7]   & --- & Halt \\
\bottomrule
\end{tabular}%
}
\caption{Failure recovery ($n{=}120$, failure rates 0.1--0.3; Wilson intervals~\citep{wilson1927}). \textbf{Wall} is mean wall-clock on the real-inference K=8, L=8 grid. \textbf{Four systems reach full recovery}; among them the differentiator is wall-clock, not recovery rate: Agent-Kernel is 3.3--12.7$\times$ faster than B3 and ${\sim}$10$\times$ faster than Sherlock-generous across grids (App.~\ref{app:sherlock},~\ref{app:b3}). McNemar $p{<}0.0001$ separates the halt/whole-replay group. \textbf{Budgets:} Agent-Kernel retries each failed node with budget 3; ReAct+retry replays the whole trajectory with budget 3; B3 retries the failing step in place. $^{\S}$Steelmanned Sherlock rollback. $^\dagger$Violates dependency ordering (fastest but 0\% dependency-correct, §\ref{sec:rq2}).}
\label{tab:failure-recovery}
\end{table}

\textbf{SGR achieves full recovery within budget=3 where full replay exhausts.} The 45pp gap between ReAct+retry (55\%) and Agent-Kernel (100\%) isolates \emph{where} you retry, not \emph{whether} you retry. Both systems have budget=3 and identical transient failure injection, but the budgets have different scopes: Agent-Kernel's is per failed node, ReAct+retry's is per trajectory. Two controls separate scope from attempt count: FullReplan gives the Agent-Kernel substrate the same 3 whole-task restarts and reaches only 56.2\% (\S\ref{sec:rq3}), while B3 gives a serial ReAct per-step scope and recovers fully. On a depth-8 DAG, whole-trajectory replay re-executes 7 completed layers before reaching the failure; any new transient failure in those layers decrements the budget without progress. SGR skips completed layers entirely, spending budget \emph{only} on the actual failure point and its downstream dependents. AsyncLM also achieves 100\% recovery (Table~\ref{tab:failure-recovery}), but at the cost of dependency correctness: it dispatches all nodes simultaneously, producing 0\% scenario-level correctness on ordered workloads (§\ref{sec:rq2}). The other full-recoverers (B3, Sherlock-generous) are dependency-correct only because they run serially; Agent-Kernel alone preserves dependency correctness \emph{while} exploiting parallelism (next paragraph).

\textbf{SGR is not the only mechanism that recovers, but it is the fastest.} Two stronger comparators in Table~\ref{tab:failure-recovery} also reach 100\% recovery, so the differentiator is wall-clock. \emph{ReAct-per-step-retry} (B3) recovers fully but runs serially, so SGR is 3.3--12.7$\times$ faster (App.~\ref{app:b3}). Our \emph{Sherlock-generous} adaptation is ${\sim}$10$\times$ slower from two costs: serial transitive-closure replay \emph{and} an always-on verifier pass on every intermediate output (its wall-clock is roughly flat at 139--167s even at failure rate 0, App.~\ref{app:sherlock}), so this gap reflects our adaptation, not necessarily an optimally-engineered Sherlock. The claim we defend is that SGR delivers exception-triggered, zero-shot recovery without a verifier or planner install.

\textbf{Efficiency implication.} Because SGR re-executes only the failed subgraph, it avoids re-generating completed siblings' outputs: on a K=8, L=8 DAG with one failure at depth 4, FullReplan re-executes 32 completed nodes per attempt while SGR touches only the failed node and its 4 descendants, and the token savings compound with DAG size (Table~\ref{tab:efficiency}).

\textbf{Cross-backbone stability.} Recovery generalizes across the Claude family: Agent-Kernel reaches 100\% on Sonnet 4.5 and Haiku 4.5 at K$\in\{$8,32$\}$ and on Opus 4.5 at K=8 (all fr=0.3), while the ReAct baselines recover partially only on Opus 4.5 (App.~\ref{app:b3}). SGR's effectiveness is not backbone-specific.

\subsection{Wall-Clock and Correctness on Real Workloads}\label{sec:rq2}

\begin{table}[t]
\centering
\footnotesize
\setlength{\tabcolsep}{2.4pt}
\begin{tabular}{@{}l rrr r l@{}}
\toprule
Benchmark & AK & ReAct & Speed & Acc. & Structure \\
\midrule
AutomationBench & 5.5s & 24.8s & \textbf{4.5$\times$} & 5\% & Indep.\ APIs \\
InterruptBench & 14.2s & 38.1s & \textbf{2.7$\times$} & 44\% & Non-block. \\
AAR Hard & 72.4s & 85.2s & 1.18$\times$ & 39\% & Diamond \\
Synthetic K=32 & 22.1s & 281.4s & \textbf{12.7$\times$} & -- & 8$\times$32 DAG \\
\midrule
tau-bench & 17.2s & 17.1s & 1.0$\times$ & 84\% & Linear \\
Gaia2 & 18.2s & 18.0s & 1.0$\times$ & 39\% & Linear \\
\bottomrule
\end{tabular}
\caption{Real LLM inference results (task counts: AutomationBench 20, InterruptBench 113, AAR Hard 15, tau-bench 165, Gaia2 640). \textbf{Acc.} is task solve-rate; AK and ReAct match within noise (same backbone), with rare scheduling-induced divergences (e.g., one diverged task on tau-bench retail, App.~\ref{app:b3}). Top: parallel workloads where AK delivers speedup. Bottom: standard linear-chain benchmarks. Execution-completion rate (a separate reliability metric on which AK trails by 1--2\% due to tail-latency timeouts) is in App.~\ref{app:taubench}. Wall-clock cells are 3-seed means (point estimates; per-seed dispersion not characterized).}
\label{tab:external}
\end{table}

\textbf{Gains scale with DAG parallelism.} AutomationBench (cross-application REST API orchestration) has naturally independent service calls, and AK forks them in parallel for a 4.5$\times$ speedup. InterruptBench demonstrates the non-blocking main loop: main-loop acknowledgement latency (MLAL, the time for the loop to accept new work while sub-agents run) is 0.15\,ms for Agent-Kernel versus 1{,}600--2{,}000\,ms for the blocking baselines, because new \texttt{fork} calls execute immediately while background sub-agents continue (App.~\ref{app:interrupt}).

\textbf{Task accuracy on standard benchmarks.} On tau-bench~\citep{yao2024taubench} (1,320 runs) and Gaia2~\citep{froger2026gaia2} (2,560 runs), Agent-Kernel achieves 84\% and 39\% task accuracy, matching ReAct-serial within noise on these linear workloads (B3 tracks ReAct-serial here, having no fan-out to exploit; its gap to SGR appears only on parallel failure grids, App.~\ref{app:b3}). The substrate thus adds recovery and parallelism without degrading accuracy. We avoid the stronger phrase ``zero overhead,'' since some long linear chains show small shortfalls (App.~\ref{app:taubench}) and context inheritance carries a fan-in-dependent token cost (below).

\textbf{Scaling by K.} On failure-free scenarios, Agent-Kernel's latency is roughly flat above K=2 (DSP saturates the 3-region API pool) while ReAct grows linearly (K=32: 22.1s vs 281.4s). AK trails AsyncLM below K$\approx$8 (scheduling overhead) then overtakes it, as DSP advances layers without waiting for slow outliers (App.~\ref{app:wallclock}).

\begin{table}[t]
\centering
\small
\setlength{\tabcolsep}{4pt}
\begin{tabular}{@{}l cc cc cc@{}}
\toprule
& \multicolumn{2}{c}{Raw-concat} & \multicolumn{2}{c}{Summarizer} & \multicolumn{2}{c}{Top-$k$} \\
\cmidrule(lr){2-3} \cmidrule(lr){4-5} \cmidrule(lr){6-7}
$K$ & Acc & Tok & Acc & Tok & Acc & Tok \\
\midrule
1   & 100\% & 86    & 100\% & 80    & 100\% & 86  \\
16  & 100\% & 339   & 100\% & 243   & 100\% & \textbf{136} \\
32  & 100\% & 608   & 100\% & 416   & 100\% & \textbf{136} \\
128 & 100\% & 2{,}363 & 100\% & 1{,}588 & 100\% & \textbf{140} \\
\bottomrule
\end{tabular}
\caption{Context-inheritance $K$-sweep (Sonnet 4.5). Acc = task accuracy; Tok = mean injected-context tokens/child prompt. Accuracy is flat at 100\% to $K{=}128$; top-$k$ retrieval holds token cost roughly constant (17$\times$ cheaper than raw-concat at $K{=}128$). Full grid ($K{\in}\{1..128\}$, 3 seeds) in App.~\ref{app:ksweep}.}
\label{tab:ksweep-main}
\end{table}

\textbf{Context inheritance does not degrade accuracy at scale.} A concern is that as K grows, DSP injects more parent output into each child prompt, risking context saturation. We swept K$\in\{$1..128$\}$ under three context policies (Table~\ref{tab:ksweep-main}): accuracy stays flat at 100\% through K=128 on Sonnet 4.5 (Haiku 4.5 confirms at K$\in\{$1,4,16,32$\}$). This is a ceiling result (100\% even at K=1), so it rules out saturation only in this easy difficulty band, not in general. The effect that does appear is token cost: top-$k$ retrieval is 17$\times$ cheaper than raw concatenation at K=128, so we recommend top-$k$ retrieval as the default for K$\geq$16 (App.~\ref{app:ksweep}).

\textbf{Dependency correctness.} On a hash-chain benchmark where children cryptographically verify parent outputs (1,440 runs), Agent-Kernel achieves 100\% scenario-level correctness (Clopper--Pearson 95\% CI~\citep{clopper1934} [99.7\%, 100.0\%]). AsyncLM, which dispatches all subtasks at $t_0$ ignoring \texttt{depends\_on} edges, collapses to 0\% [0.0\%, 0.3\%]. This is structural by construction, but quantifies a correctness property that prior work left qualitative.

\subsection{Ablation: Is SGR the Key Design Choice?}\label{sec:rq3}

\begin{table}[t]
\centering
\small
\begin{tabular}{l r}
\toprule
Variant & Overall success \\
\midrule
Agent-Kernel (full) & \textbf{100.0\%} \\
NoSAB & 100.0\% \\
NoDSP & 100.0\% \\
FullReplan (matched budget) & 56.2\% \\
\textbf{NoSGR} & \textbf{39.4\%} \\
\bottomrule
\end{tabular}
\caption{Ablation: overall success across the failure grid. Removing SGR collapses recovery to the halt-on-failure level (39.4\%); FullReplan (matched budget, whole-task restart) reaches only 56.2\%; removing SAB or DSP is invisible on this benchmark. Per-$f_r$ breakdown in App.~\ref{app:ablation}.}
\label{tab:ablation-compact}
\end{table}

\textbf{Yes. SGR is the only primitive whose removal degrades performance} (Table~\ref{tab:ablation-compact}). Agent-Kernel (full), NoSAB, and NoDSP all reach 100\%, but \textbf{NoSGR collapses to 39.4\%} (the halt-on-failure level), and \textbf{FullReplan} (same budget, whole-task restart) reaches only 56.2\%. This confirms \emph{retry alone} is insufficient; \emph{where you retry} is the design choice that matters. SAB and DSP are enabling substrate (durability, reactive scheduling) whose value emerges under conditions outside this benchmark (process crashes, K$>$32 fan-out).

\begin{table}[t]
\centering
\small
\begin{tabular}{@{}l c c c@{}}
\toprule
System (real binary) & Recovery & Re-exec. & Wall \\
\midrule
\textbf{Agent-Kernel (SGR)} & \textbf{60/60} & \textbf{1.0} & \textbf{0.12s} \\
LangGraph 1.2$^{\ast}$ & 60/60 & 1.0 & 0.52s \\
LlamaIndex 0.14 & 0/60 & 0.0 & n/a \\
AutoGen v0.4 & 0/60 & 0.0 & n/a \\
\bottomrule
\end{tabular}
\caption{\textbf{Real-binary integration} on a diamond fan-out DAG with one failing leaf (12 configs $\times$ 5 seeds $=$ 60 cells/system). Re-exec.\ = cells re-executed on recovery. $^{\ast}$LangGraph recovers only with a \emph{custom} \texttt{retry\_on} filter; its default excludes our injected exceptions, and AutoGen~v0.4 has no default runtime retry (App.~\ref{app:integration}).}
\label{tab:integration-main}
\end{table}

\textbf{Framework default recovery paths lack automatic selective retry.} On the same grid, \emph{primitive re-implementations} of the default scheduling strategy of LangGraph~\citep{chase2024langgraph}, AutoGen~\citep{wu2023autogen}, and LlamaIndex~\citep{liu2022llamaindex} collapse to ${\sim}$39\% under failure, matching non-retrying baselines, because each halts or requires full state rewind when a node fails (full table in App.~\ref{app:frameworks}). These are re-implementations of the scheduling primitive, not runs of the shipped binaries. Running the actual binaries confirms the boundary (Table~\ref{tab:integration-main}): with a \emph{custom} retry filter LangGraph~1.2 recovers fully, but its \emph{default} filter excludes the application-level exceptions we inject, and AutoGen~v0.4 has no default runtime retry primitive. Agent-Kernel is therefore the only system that resets transitive descendants and preserves siblings without user configuration.

\textbf{Cross-backbone replication.} The pattern holds across the Claude family (Table~\ref{tab:crossbackbone-main}): Agent-Kernel reaches 100\% recovery on Sonnet 4.5, Haiku 4.5, and Opus 4.5, while the ReAct baselines recover partially only on Opus 4.5.

\begin{table}[t]
\centering
\small
\begin{tabular}{@{}l l c c@{}}
\toprule
Backbone & Config & AK & ReAct \\
\midrule
Sonnet 4.5 & K=8, fr=0.3  & \textbf{100\%} & 0\% \\
Sonnet 4.5 & K=32, fr=0.3 & \textbf{100\%} & 0\% \\
Haiku 4.5  & K=8, fr=0.3  & \textbf{100\%} & 0\% \\
Haiku 4.5  & K=32, fr=0.3 & \textbf{100\%} & 0\% \\
Opus 4.5   & K=8, fr=0.3  & \textbf{100\%} & 50\% \\
\bottomrule
\end{tabular}
\caption{Cross-backbone recovery across the Claude family (transient failures). The ReAct column gives ReAct-serial; ReAct+retry matches it on every cell.}
\label{tab:crossbackbone-main}
\end{table}

\textbf{Token efficiency under failure.} By avoiding re-generation of completed siblings, SGR generates an estimated $\sim$2.7$\times$ fewer tokens than FullReplan ($\sim$35K vs $\sim$96K) at 2.2$\times$ higher success on a K=8, L=8 DAG at fr=0.3 (App.~\ref{app:efficiency}); the gap compounds with scale.

\textbf{Auto-Fork.} An extension automatically decomposes a sequential task into a parallel DAG via one LLM decomposition call (cycle-validated); on AAR-Linear it finds latent parallelism for a 3.3$\times$ speedup while inheriting SGR's recovery guarantees (App.~\ref{app:autofork}).

\section{Discussion}\label{sec:discussion}

\textbf{Where you retry, and on what plane.} The 44pp gap between SGR (100\%) and FullReplan (56.2\%) under identical budgets (§\ref{sec:rq3}) isolates retry \emph{scope}, not retry itself, as the key variable: any multi-step LLM recovery mechanism should preserve completed work rather than replay from the root. But that pattern is established (Spark, Erlang/OTP, Dryad), and recent runtimes (Sherlock, Atomix, AgentGit) already preserve completed work; what none touch is the LLM content plane: composing upstream text into a downstream prompt (context inheritance) and rewriting a prompt on format drift (prompt-mutation retry). Our ablations (App.~\ref{app:contentplane}) show each is necessary, by construction, in temperature-0 probes rather than large-$N$ benchmarks.

\textbf{What is new for LLM agents.} Each ingredient (selective DAG recovery, persistence, context passing, error-feedback retry) is established, and we cite its lineage. The contribution is the system result built from them: (i)~the measured finding that retry \emph{scope}, not retry itself, is the lever (SGR 100\% vs.\ FullReplan 56.2\% at the same budget); (ii)~recovery-time context inheritance as a first-class policy, which the ablation shows is necessary; (iii)~a formal state-preservation guarantee, with dependency correctness quantified (100\% vs.\ 0\% for dependency-blind dispatch); and (iv)~recovery without configuration, which the shipped LangGraph, AutoGen, and LlamaIndex binaries do not provide by default.

\textbf{Detection versus recovery.} A verifier such as Sherlock's catches semantic errors that an exception cannot, so it is the stronger \emph{detector}; SGR is a \emph{recovery} substrate, and the two compose rather than compete (\S\ref{sec:sgr}). Our ${\sim}10\times$ speed gap over Sherlock-generous is therefore not a claim about detection quality: it comes from that adaptation's serial transitive-closure replay and its verifier pass on every intermediate output (App.~\ref{app:sherlock}). What we defend is narrower: recovery itself needs no verifier or planner install, and when semantic detection is needed, a verifier's signal can drive the same reset. The open frontier is therefore \emph{semantic} failures. In a preliminary study, a schema check plus prompt-mutation retry recovers 14/18 format-drift scenarios vs.\ 6/18 for blind retry (App.~\ref{app:pmr-ablation}), so subgraph reset appears to compose with language-aware detection and recovery. Testing that at scale is future work.

\section{Conclusion}\label{sec:conclusion}

We presented Agent-Kernel, a non-blocking runtime that ports the established subgraph-selective recovery pattern to the LLM content plane. Its primitive, Sub-Graph Recovery (SGR), resets a failed node's transitive descendants while preserving completed siblings and composing each parent's text into its children's prompts (context inheritance), on a durable substrate with a formal guarantee. Across 14,434 runs it achieves full recovery where whole-trajectory replay exhausts (100\% vs 55\%). Against baselines that also recover fully, its advantage is wall-clock (3.3--12.7$\times$). The pattern is established; with content-plane semantics it becomes a practical reliability primitive for agentic systems.

\clearpage
\section*{Limitations}

While Agent-Kernel demonstrates strong performance on failure recovery and parallel orchestration, six limitations remain.

First, our main failure model evaluates only transient crashes that succeed on retry. The exception is the trigger we control, not a limit of the mechanism, since any detector that marks a node failed fires the same reset (\S\ref{sec:sgr}), but we evaluate other detectors only in small probes. Real LLM agent failures include hallucination, format drift, and capability limits where blind re-execution is insufficient. SGR's architecture supports modified retry with error-context injection, and our preliminary study on semantic failures is encouraging (14/18 vs 6/18 for blind retry, \S\ref{sec:discussion}). A complete large-$N$ evaluation across semantic failure patterns remains future work. Extending SGR to handle non-transient failures would require integrating output validation into the retry loop: determining \emph{when} a completed output is incorrect and should trigger subgraph reset.

Second, Agent-Kernel's advantages are conditional on workload structure. On linear chains without exploitable parallelism or failure injection (tau-bench, Gaia2), the system ties ReAct-serial within noise, with negligible overhead and no benefit. Users whose workloads are exclusively sequential will see no improvement from adopting Agent-Kernel.

Third, our grid-scale framework comparison uses primitive re-implementations of each framework's default scheduling strategy rather than the full binaries. The real-binary integration tests for LangGraph~1.2, LlamaIndex~0.14, and AutoGen~v0.4 (Table~\ref{tab:integration-main}, App.~\ref{app:integration}) cover a single diamond fan-out topology; broader integration across topologies and framework configurations would further strengthen the comparison.

Fourth, results cover the Claude family (Sonnet 4.5, Haiku 4.5, Opus 4.5). Cross-provider validation (GPT-class and open-source models) is absent, and the system's reliance on structured output parsing means backbone changes may require parser adaptation.

Fifth, the state-preservation guarantee covers the state SAB records, not external side effects (\S\ref{sec:sgr}). We do not evaluate tools with non-idempotent effects, idempotency keys, or compensation: a node that fails after a side-effecting call can repeat that effect when it is retried. Composing SGR with a transactional tool boundary such as Atomix~\citep{mohammadi2026atomix}, or with application-specific idempotency and compensation logic for retryable failure modes, is future work.

Finally, our positive results on context inheritance are bounded by the backbones and task distribution we test. The K-sweep shows accuracy is flat to K=128 on Sonnet 4.5, and to K=32 on Haiku 4.5 (App.~\ref{app:ksweep}), so saturation does not appear here. However, longer or more heterogeneous intermediate outputs, or weaker backbones, could surface degradation that our top-$k$ retrieval policy only partially mitigates.

\section*{Ethics Statement}

This work introduces runtime infrastructure for LLM agent orchestration. We do not train new models, collect personal data, or deploy the system in sensitive domains. All experiments use commercially available LLM API endpoints under standard terms of service. The synthetic benchmarks contain no harmful content. We acknowledge that more capable agent orchestration could amplify both beneficial and harmful uses of LLM agents; responsible deployment requires appropriate access controls and monitoring at the application layer, which is outside our scope.

Throughout the development of this work, we used generative AI systems for language refinement and code test generation. All core research contributions, system design, experimental methodology, and analysis are the result of human intellectual effort.

\bibliographystyle{acl_natbib}
\bibliography{references}

\begin{thebibliography}{34}
\providecommand{\natexlab}[1]{#1}

\bibitem[{{Apache Software Foundation}(2015)}]{airflow2015}
{Apache Software Foundation}. 2015.
\newblock {Apache Airflow}.
\newblock \url{https://airflow.apache.org/}.
\newblock Software.

\bibitem[{Armstrong(2003)}]{armstrong2003erlang}
Joe Armstrong. 2003.
\newblock \emph{Making Reliable Distributed Systems in the Presence of Software
  Errors}.
\newblock Ph.D. thesis, Royal Institute of Technology (KTH), Stockholm.

\bibitem[{Babuji et~al.(2019)Babuji, Woodard, Li, Katz, Clifford, Kumar,
  Lacinski, Chard, Wozniak, Foster, Wilde, and Chard}]{parsl2019}
Yadu Babuji, Anna Woodard, Zhuozhao Li, Daniel~S. Katz, Ben Clifford, Rohan
  Kumar, Lukasz Lacinski, Ryan Chard, Justin~M. Wozniak, Ian Foster, Michael
  Wilde, and Kyle Chard. 2019.
\newblock \href {https://doi.org/10.1145/3307681.3325400} {{Parsl}: Pervasive
  parallel programming in {Python}}.
\newblock In \emph{Proceedings of the 28th International Symposium on
  High-Performance Parallel and Distributed Computing}, pages 25--36.
  Association for Computing Machinery.

\bibitem[{Burckhardt et~al.(2021)Burckhardt, Gillum, Justo, Kallas, McMahon,
  and Meiklejohn}]{burckhardt2021durable}
Sebastian Burckhardt, Chris Gillum, David Justo, Konstantinos Kallas, Connor
  McMahon, and Christopher~S. Meiklejohn. 2021.
\newblock \href {https://doi.org/10.1145/3485510} {{Durable Functions}:
  Semantics for stateful serverless}.
\newblock \emph{Proceedings of the ACM on Programming Languages},
  5(OOPSLA):1--27.

\bibitem[{Clopper and Pearson(1934)}]{clopper1934}
Charles~J. Clopper and Egon~S. Pearson. 1934.
\newblock \href {https://doi.org/10.1093/biomet/26.4.404} {The use of
  confidence or fiducial limits illustrated in the case of the binomial}.
\newblock \emph{Biometrika}, 26(4):404--413.

\bibitem[{Dichev et~al.(2017)Dichev, Jordan, Tovletoglou, Heller, Nikolopoulos,
  Karakonstantis, and Gillan}]{dichev2017rollback}
Kiril Dichev, Herbert Jordan, Konstantinos Tovletoglou, Thomas Heller,
  Dimitrios~S. Nikolopoulos, Georgios Karakonstantis, and Charles Gillan. 2017.
\newblock \href {https://doi.org/10.48550/arXiv.1705.10208} {Dependency-aware
  rollback and checkpoint-restart for distributed task-based runtimes}.
\newblock \emph{arXiv preprint arXiv:1705.10208}.

\bibitem[{Elnozahy et~al.(2002)Elnozahy, Alvisi, Wang, and
  Johnson}]{elnozahy2002survey}
Elmootazbellah~N. Elnozahy, Lorenzo Alvisi, Yi-Min Wang, and David~B. Johnson.
  2002.
\newblock \href {https://doi.org/10.1145/568522.568525} {A survey of
  rollback-recovery protocols in message-passing systems}.
\newblock \emph{ACM Computing Surveys}, 34(3):375--408.

\bibitem[{Froger et~al.(2026)Froger, Andrews, Bettini, Budhiraja, Cabral, Do,
  Garreau, Gaya, Lauren{\c{c}}on, Lecanu, Malkan, Mekala, M{\'e}nard,
  Moreno-Torres~Bertran, Piterbarg, Plekhanov, Rita, Rusakov, Vorotilov, Wang,
  Yu, Benhalloum, Mialon, and Scialom}]{froger2026gaia2}
Romain Froger, Pierre Andrews, Matteo Bettini, Amar Budhiraja, Ricardo~Silveira
  Cabral, Virginie Do, Emilien Garreau, Jean-Baptiste Gaya, Hugo
  Lauren{\c{c}}on, Maxime Lecanu, Kunal Malkan, Dheeraj Mekala, Pierre
  M{\'e}nard, Gerard Moreno-Torres~Bertran, Ulyana Piterbarg, Mikhail
  Plekhanov, Mathieu Rita, Andrey Rusakov, Vladislav Vorotilov, and 5 others.
  2026.
\newblock \href {https://openreview.net/forum?id=9gw03JpKK4} {{Gaia2}:
  Benchmarking {LLM} agents on dynamic and asynchronous environments}.
\newblock In \emph{The Fourteenth International Conference on Learning
  Representations}.

\bibitem[{Gim et~al.(2024)Gim, Lee, and Zhong}]{gim2024asynclm}
In~Gim, Seung-seob Lee, and Lin Zhong. 2024.
\newblock \href {https://doi.org/10.48550/arXiv.2412.07017} {Asynchronous {LLM}
  function calling}.
\newblock \emph{arXiv preprint arXiv:2412.07017}.

\bibitem[{Isard et~al.(2007)Isard, Budiu, Yu, Birrell, and
  Fetterly}]{isard2007dryad}
Michael Isard, Mihai Budiu, Yuan Yu, Andrew Birrell, and Dennis Fetterly. 2007.
\newblock \href {https://doi.org/10.1145/1272996.1273005} {{Dryad}: Distributed
  data-parallel programs from sequential building blocks}.
\newblock In \emph{Proceedings of the 2nd ACM SIGOPS/EuroSys European
  Conference on Computer Systems (EuroSys '07)}, pages 59--72.

\bibitem[{Kim et~al.(2024)Kim, Moon, Tabrizi, Lee, Mahoney, Keutzer, and
  Gholami}]{kim2024llmcompiler}
Sehoon Kim, Suhong Moon, Ryan Tabrizi, Nicholas Lee, Michael~W. Mahoney, Kurt
  Keutzer, and Amir Gholami. 2024.
\newblock \href {https://proceedings.mlr.press/v235/kim24y.html} {An {LLM}
  compiler for parallel function calling}.
\newblock In \emph{Proceedings of the 41st International Conference on Machine
  Learning}, volume 235 of \emph{Proceedings of Machine Learning Research},
  pages 24370--24391. PMLR.

\bibitem[{Kim et~al.(2026)Kim, Lee, Kim, Raheja, and Kang}]{aar2026}
Zae~Myung Kim, Dongseok Lee, Jaehyung Kim, Vipul Raheja, and Dongyeop Kang.
  2026.
\newblock {The Amazing Agent Race}: Strong tool users, weak navigators.
\newblock \emph{arXiv preprint arXiv:2604.10261}.

\bibitem[{{LangChain}(2024)}]{chase2024langgraph}
{LangChain}. 2024.
\newblock {LangGraph}: Building stateful, multi-actor applications with {LLM}s.
\newblock \url{https://github.com/langchain-ai/langgraph}.
\newblock Software library. Accessed 2026-05-25.

\bibitem[{Li et~al.(2025)Li, Ping, Chen, Qi, Wang, Luo, and
  Zhang}]{li2025agentgit}
Yang Li, Siqi Ping, Xiyu Chen, Xiaojian Qi, Zigan Wang, Ye~Luo, and Xiaowei
  Zhang. 2025.
\newblock {AgentGit}: A version control framework for reliable and scalable
  {LLM}-powered multi-agent systems.
\newblock \emph{arXiv preprint arXiv:2511.00628}.

\bibitem[{Liu(2022)}]{liu2022llamaindex}
Jerry Liu. 2022.
\newblock {LlamaIndex}.
\newblock \url{https://github.com/run-llama/llama_index}.
\newblock Software library.

\bibitem[{Madaan et~al.(2023)Madaan, Tandon, Gupta, Hallinan, Gao, Wiegreffe,
  Alon, Dziri, Prabhumoye, Yang, Gupta, Majumder, Hermann, Welleck,
  Yazdanbakhsh, and Clark}]{madaan2023selfrefine}
Aman Madaan, Niket Tandon, Prakhar Gupta, Skyler Hallinan, Luyu Gao, Sarah
  Wiegreffe, Uri Alon, Nouha Dziri, Shrimai Prabhumoye, Yiming Yang, Shashank
  Gupta, Bodhisattwa~Prasad Majumder, Katherine Hermann, Sean Welleck, Amir
  Yazdanbakhsh, and Peter Clark. 2023.
\newblock \href
  {https://proceedings.neurips.cc/paper_files/paper/2023/hash/91edff07232fb1b55a505a9e9f6c0ff3-Abstract-Conference.html}
  {{Self-Refine}: Iterative refinement with self-feedback}.
\newblock In \emph{Advances in Neural Information Processing Systems},
  volume~36, pages 46534--46594. Curran Associates, Inc.

\bibitem[{Mei et~al.(2025)Mei, Zhu, Xu, Jin, Hua, Li, Xu, Ye, Ge, and
  Zhang}]{mei2024aios}
Kai Mei, Xi~Zhu, Wujiang Xu, Mingyu Jin, Wenyue Hua, Zelong Li, Shuyuan Xu,
  Ruosong Ye, Yingqiang Ge, and Yongfeng Zhang. 2025.
\newblock \href {https://openreview.net/forum?id=L4HHkCDz2x} {{AIOS}: {LLM}
  agent operating system}.
\newblock In \emph{Second Conference on Language Modeling}.

\bibitem[{Mohammadi et~al.(2026)Mohammadi, Potamitis, Klein, Arora, and
  Bindschaedler}]{mohammadi2026atomix}
Bardia Mohammadi, Nearchos Potamitis, Lars Klein, Akhil Arora, and Laurent
  Bindschaedler. 2026.
\newblock {Atomix}: Timely, transactional tool use for reliable agentic
  workflows.
\newblock \emph{arXiv preprint arXiv:2602.14849}.

\bibitem[{{Prefect Technologies}(2019)}]{prefect2018}
{Prefect Technologies}. 2019.
\newblock {Prefect}: The dataflow automation platform.
\newblock \url{https://www.prefect.io/}.
\newblock Software.

\bibitem[{Ro et~al.(2025)Ro, Qiu, Goiri, Fonseca, Bianchini, Akella, Wang,
  Erez, and Choukse}]{ro2025sherlock}
Yeonju Ro, Haoran Qiu, {\'I}{\~n}igo Goiri, Rodrigo Fonseca, Ricardo Bianchini,
  Aditya Akella, Zhangyang Wang, Mattan Erez, and Esha Choukse. 2025.
\newblock {Sherlock}: Reliable and efficient agentic workflow execution.
\newblock \emph{arXiv preprint arXiv:2511.00330}.

\bibitem[{Shepard and Salimans(2026)}]{automationbench2026}
Daniel Shepard and Robin Salimans. 2026.
\newblock {AutomationBench}.
\newblock \emph{arXiv preprint arXiv:2604.18934}.

\bibitem[{Shinn et~al.(2023)Shinn, Cassano, Gopinath, Narasimhan, and
  Yao}]{shinn2023reflexion}
Noah Shinn, Federico Cassano, Ashwin Gopinath, Karthik Narasimhan, and Shunyu
  Yao. 2023.
\newblock \href
  {https://proceedings.neurips.cc/paper_files/paper/2023/hash/1b44b878bb782e6954cd888628510e90-Abstract-Conference.html}
  {Reflexion: Language agents with verbal reinforcement learning}.
\newblock In \emph{Advances in Neural Information Processing Systems},
  volume~36, pages 8634--8652. Curran Associates, Inc.

\bibitem[{Strom and Yemini(1985)}]{strom1985optimistic}
Robert~E. Strom and Shaula Yemini. 1985.
\newblock \href {https://doi.org/10.1145/3959.3962} {Optimistic recovery in
  distributed systems}.
\newblock \emph{ACM Transactions on Computer Systems}, 3(3):204--226.

\bibitem[{{Temporal Technologies}(2020)}]{temporal2020}
{Temporal Technologies}. 2020.
\newblock {Temporal}: Durable execution platform.
\newblock \url{https://temporal.io/}.
\newblock Software.

\bibitem[{Thompson(2025)}]{dualstate2025}
Matthew Thompson. 2025.
\newblock The dual-state architecture for reliable {LLM} agents.
\newblock \emph{arXiv preprint arXiv:2512.20660}.

\bibitem[{Trivedi et~al.(2024)Trivedi, Khot, Hartmann, Manku, Dong, Li, Gupta,
  Sabharwal, and Balasubramanian}]{trivedi2024appworld}
Harsh Trivedi, Tushar Khot, Mareike Hartmann, Ruskin Manku, Vinty Dong, Edward
  Li, Shashank Gupta, Ashish Sabharwal, and Niranjan Balasubramanian. 2024.
\newblock \href {https://doi.org/10.18653/v1/2024.acl-long.850} {{AppWorld}: A
  controllable world of apps and people for benchmarking interactive coding
  agents}.
\newblock In \emph{Proceedings of the 62nd Annual Meeting of the Association
  for Computational Linguistics (Volume 1: Long Papers)}, pages 16022--16076,
  Bangkok, Thailand. Association for Computational Linguistics.

\bibitem[{Wilson(1927)}]{wilson1927}
Edwin~B. Wilson. 1927.
\newblock \href {https://doi.org/10.1080/01621459.1927.10502953} {Probable
  inference, the law of succession, and statistical inference}.
\newblock \emph{Journal of the American Statistical Association},
  22(158):209--212.

\bibitem[{Wu et~al.(2024)Wu, Bansal, Zhang, Wu, Li, Zhu, Jiang, Zhang, Zhang,
  Liu, Awadallah, White, Burger, and Wang}]{wu2023autogen}
Qingyun Wu, Gagan Bansal, Jieyu Zhang, Yiran Wu, Beibin Li, Erkang Zhu,
  Li~Jiang, Xiaoyun Zhang, Shaokun Zhang, Jiale Liu, Ahmed~Hassan Awadallah,
  Ryen~W. White, Doug Burger, and Chi Wang. 2024.
\newblock \href {https://openreview.net/forum?id=BAakY1hNKS} {{AutoGen}:
  Enabling next-gen {LLM} applications via multi-agent conversations}.
\newblock In \emph{First Conference on Language Modeling}.

\bibitem[{Yao et~al.(2025)Yao, Shinn, Razavi, and Narasimhan}]{yao2024taubench}
Shunyu Yao, Noah Shinn, Pedram Razavi, and Karthik Narasimhan. 2025.
\newblock \href {https://openreview.net/forum?id=roNSXZpUDN} {$\tau$-bench: A
  benchmark for tool-agent-user interaction in real-world domains}.
\newblock In \emph{The Thirteenth International Conference on Learning
  Representations}.

\bibitem[{Yao et~al.(2023)Yao, Zhao, Yu, Du, Shafran, Narasimhan, and
  Cao}]{yao2023react}
Shunyu Yao, Jeffrey Zhao, Dian Yu, Nan Du, Izhak Shafran, Karthik~R.
  Narasimhan, and Yuan Cao. 2023.
\newblock \href {https://openreview.net/forum?id=WE_vluYUL-X} {{ReAct}:
  Synergizing reasoning and acting in language models}.
\newblock In \emph{The Eleventh International Conference on Learning
  Representations}.

\bibitem[{Yu(2026)}]{adaptorch2026}
Geunbin Yu. 2026.
\newblock {AdaptOrch}: Task-adaptive multi-agent orchestration in the era of
  {LLM} performance convergence.
\newblock \emph{arXiv preprint arXiv:2602.16873}.

\bibitem[{Yu et~al.(2025)Yu, Ding, and Sato}]{dyntaskmas2025}
Junwei Yu, Yepeng Ding, and Hiroyuki Sato. 2025.
\newblock \href {https://doi.org/10.1609/icaps.v35i1.36130} {{DynTaskMAS}: A
  dynamic task graph-driven framework for asynchronous and parallel {LLM}-based
  multi-agent systems}.
\newblock In \emph{Proceedings of the International Conference on Automated
  Planning and Scheduling}, volume~35, pages 288--296.

\bibitem[{Zaharia et~al.(2012)Zaharia, Chowdhury, Das, Dave, Ma, McCauley,
  Franklin, Shenker, and Stoica}]{zaharia2012rdd}
Matei Zaharia, Mosharaf Chowdhury, Tathagata Das, Ankur Dave, Justin Ma, Murphy
  McCauley, Michael~J. Franklin, Scott Shenker, and Ion Stoica. 2012.
\newblock \href
  {https://www.usenix.org/conference/nsdi12/technical-sessions/presentation/zaharia}
  {Resilient distributed datasets: A fault-tolerant abstraction for in-memory
  cluster computing}.
\newblock In \emph{9th USENIX Symposium on Networked Systems Design and
  Implementation (NSDI 12)}, pages 15--28, San Jose, CA. USENIX Association.

\bibitem[{Zou et~al.(2026)Zou, Miao, Huang, Chen, Zhou, Zhang, Wu, Fang, Gu,
  Zhang, Zheng, Wang, Nian, Li, Fan, He, Zhang, Liu, and
  Yu}]{interruptbench2026}
Henry~Peng Zou, Chunyu Miao, Wei-Chieh Huang, Yankai Chen, Yue Zhou, Hanrong
  Zhang, Yaozu Wu, Liancheng Fang, Zhengyao Gu, Zhen Zhang, Kening Zheng,
  Fangxin Wang, Yi~Nian, Shanghao Li, Wenzhe Fan, Langzhou He, Weizhi Zhang,
  Xue Liu, and Philip~S. Yu. 2026.
\newblock When users change their mind: Evaluating interruptible agents in
  long-horizon web navigation.
\newblock \emph{arXiv preprint arXiv:2604.00892}.

\end{thebibliography}

\appendix
\section{Architecture Diagram}\label{app:architecture}

\begin{figure*}[h]
\centering
\includegraphics[width=0.72\textwidth]{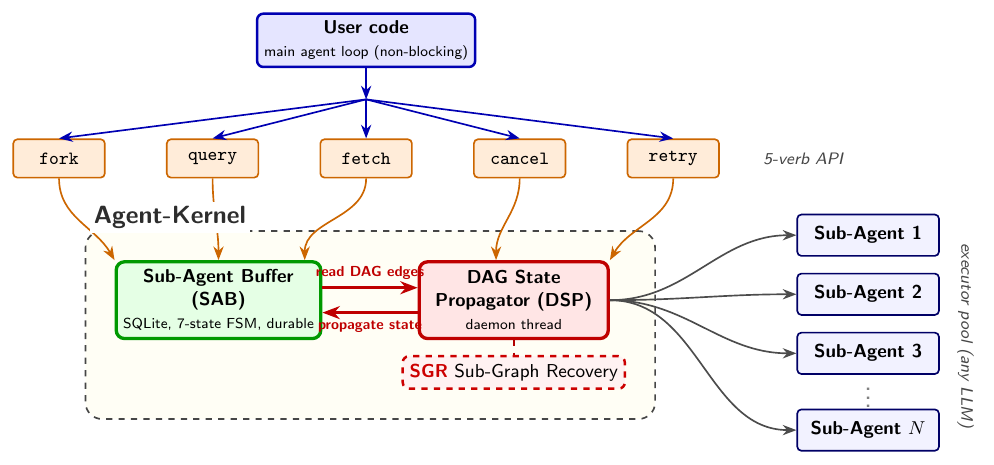}
\caption{Agent-Kernel architecture. The user calls the 5-verb API; DSP dispatches sub-agents with parents' outputs injected from SAB (context inheritance). SGR handles failure by resetting only the failed subgraph.}
\label{fig:architecture}
\end{figure*}

\section{Extended Experimental Results}\label{app:extended}

\subsection{Evaluation Footprint}\label{app:footprint}

\begin{table}[h]
\centering
\small
\setlength{\tabcolsep}{4pt}
\begin{tabular}{@{}lrr@{}}
\toprule
Evaluation & Runs & Section \\
\midrule
Synthetic (failure recovery) & 1,440 & §\ref{sec:rq1} \\
Synthetic (dependency) & 1,440 & §\ref{sec:rq2} \\
Ablation + FullReplan & 1,320 & §\ref{sec:rq3} \\
K=32 scaling (5 seeds) & 80 & §\ref{sec:rq2} \\
Framework comparison (primitive) & 360 & §\ref{sec:rq3} \\
External benchmarks & 3,289 & §\ref{sec:rq2} \\
tau-bench + Gaia2 & 3,880 & §\ref{sec:rq2} \\
\midrule
\textbf{Core subtotal} & \textbf{11,809} & \\
\midrule
Sherlock + B3 grids & 414 & App.~\ref{app:sherlock},~\ref{app:b3} \\
Cross-backbone & 135 & App.~\ref{app:b3} \\
Real-framework integration & 240 & App.~\ref{app:integration} \\
K-sweep (Sonnet+Haiku) & 1,836 & App.~\ref{app:ksweep} \\
\midrule
\textbf{Total} & \textbf{14,434} & \\
\bottomrule
\end{tabular}
\caption{Evaluation footprint. Rows above the first subtotal are the core grids and benchmarks (11,809 runs); rows below are the comparator and scaling studies (2,625 runs), for a total of 14,434 runs.}
\label{tab:eval-scale-compact}
\end{table}

\subsection{External Benchmark Results}\label{app:external}

\begin{table}[h]
\centering
\small
\resizebox{\columnwidth}{!}{%
\begin{tabular}{llcccc}
\toprule
Benchmark & Agent-Kernel & ReAct & AsyncLM & AIOS \\
\midrule
AutomationBench (20 tasks) & \textbf{5.5s} & 24.8s & 5.1s & 25.2s \\
AAR Easy (8 stops, 20 tasks) & \textbf{24.9s} & 31.2s & 6.4s & 32.3s \\
AAR Hard (25 stops, 15 tasks) & \textbf{72.4s} & 85.2s & 13.5s & 86.3s \\
InterruptBench (113 tasks) & \textbf{14.2s} & 38.1s & 9.7s & 37.8s \\
AppWorld dev (10 tasks) & 62.6s & 64.0s & 9.9s & 60.9s \\
AppWorld test\_normal (15 tasks) & 8.9s & 9.0s & 5.7s & 8.5s \\
\bottomrule
\end{tabular}%
}
\caption{Wall-clock on external benchmarks with real LLM inference. Bold marks Agent-Kernel where it is the fastest dependency-respecting system (AK/ReAct/AIOS); on the two AppWorld rows AIOS edges it out. AsyncLM ignores dependency edges, producing faster but potentially incorrect execution.}
\label{tab:external-wallclock}
\end{table}

Agent-Kernel is 4.5$\times$ faster than ReAct on AutomationBench (cross-app API orchestration with natural fork points), 2.7$\times$ on InterruptBench (non-blocking main loop), and 15--20\% on AAR (diamond patterns in dependency DAGs).

\subsection{Wall-Clock by Parallel Width K}\label{app:wallclock}

\begin{table}[h]
\centering
\small
\resizebox{\columnwidth}{!}{%
\begin{tabular}{lllll}
\toprule
K & Agent-Kernel & AsyncLM & ReAct & AIOS \\
\midrule
1 & 9.9s & 3.0s & 10.0s & 10.3s \\
4 & 9.6s & 5.1s & 29.8s & 30.6s \\
8 & 11.2s & 9.0s & 62.3s & 58.8s \\
16 & 12.2s & 15.9s & 116.3s & 118.0s \\
32 & \textbf{22.1s} & 38.9s & \textbf{281.4s} & 273.8s \\
\bottomrule
\end{tabular}%
}
\caption{Wall-clock by K (failure-free, K${\leq}$16: 3 seeds; K=32: 5 seeds). Agent-Kernel stays roughly flat while ReAct/AIOS grow linearly; AsyncLM is cheapest below K$\approx$8, after which Agent-Kernel overtakes it.}
\label{tab:wallclock}
\end{table}

\subsection{Speed-Correctness Tradeoff}\label{app:speed-correct}

AsyncLM is consistently 3--8$\times$ faster than Agent-Kernel because it submits all $N$ subtasks simultaneously regardless of dependencies. Agent-Kernel pays a wall-clock cost for correctness: it executes in DAG-layer order, with wall-clock $O(L \times t_{\text{call}})$ vs AsyncLM's $O(t_{\text{call}})$. The gap equals the dependency depth---the price of respecting execution order.

\subsection{Main-Loop Acknowledgement Latency (MLAL)}\label{app:interrupt}

We define \emph{main-loop acknowledgement latency} (MLAL) as the time for the agent's main loop to accept and begin acting on a new request while background sub-agents are still running---i.e., how quickly the loop acknowledges an interrupt, not how quickly the interrupted work finishes. On the 113-task InterruptBench slice across five systems, Agent-Kernel's MLAL is 0.15\,ms versus 1{,}600--2{,}000\,ms for the blocking baselines, because new \texttt{fork} calls return immediately while background sub-agents continue. This latency metric is distinct from the 2.7$\times$ end-to-end wall-clock speedup on the same benchmark (Table~\ref{tab:external}).

\subsection{Resource Efficiency}\label{app:resource}

\begin{table}[h]
\centering
\small
\begin{tabular}{llll}
\toprule
System & Total calls & Successes & Calls/success \\
\midrule
Agent-Kernel & ${\sim}$4,800 & 360/360 & 13.3 \\
AsyncLM & ${\sim}$2,400 & 360/360 & 6.7 \\
ReAct-serial & ${\sim}$1,200 & 142/360 & 8.5 \\
AIOS & ${\sim}$1,800 & 142/360 & 12.7 \\
\bottomrule
\end{tabular}
\caption{API calls per successful task. ReAct's low total reflects early halting, not efficiency.}
\label{tab:resource}
\end{table}

\subsection{AsyncLM Node Accuracy by K$\times$L}\label{app:asynclm-breakdown}

\begin{table}[h]
\centering
\small
\begin{tabular}{lllll}
\toprule
K & L=1 & L=2 & L=4 & L=8 \\
\midrule
1 & 50.0\% & 33.3\% & 20.0\% & 11.1\% \\
4 & 80.0\% & 44.4\% & 23.5\% & 12.1\% \\
16 & 94.1\% & 48.5\% & 24.6\% & 12.4\% \\
\bottomrule
\end{tabular}
\caption{AsyncLM node accuracy drops with depth L. Surviving accuracy equals root-node fraction.}
\label{tab:asynclm-breakdown}
\end{table}

\subsection{tau-bench and Gaia2 Full Results}\label{app:taubench}

Table~\ref{tab:taubench-full} reports wall-clock and \emph{execution-completion rate} (the fraction of runs that finish within the runner timeout without a harness-level error)---distinct from the task \emph{solve accuracy} in the main-text Table~\ref{tab:external}. Solve accuracy matches within noise for AK and ReAct (same backbone), with only rare scheduling-induced divergences (e.g., the single diverged tau-bench-retail task in App.~\ref{app:b3}); completion rate differs because AK's DAG-layer scheduling occasionally accumulates tail latency past the 60s runner timeout on long linear chains, costing 1--2\%.

The run counts in \S\ref{sec:setup} are aggregate: a \emph{run} is one (task $\times$ seed $\times$ config) cell, so the totals exceed the per-benchmark task counts in Table~\ref{tab:external}. tau-bench's 1{,}320 runs cover its 165 tasks (an $8\times$ multiplier over seeds and configs) and Gaia2's 2{,}560 runs cover its 640 tasks ($4\times$); accuracy in Table~\ref{tab:external} is reported per task, aggregated over the corresponding runs.

\begin{table}[h]
\centering
\small
\resizebox{\columnwidth}{!}{%
\begin{tabular}{lcccc}
\toprule
Benchmark (wall / completion) & AK & ReAct & AsyncLM & AIOS \\
\midrule
tau-bench (Mac) & 17.2s / 98.2\% & 17.1s / 100\% & 4.9s / 100\%* & 17.1s / 100\% \\
tau-bench (dev-dsk) & 16.2s / 99.4\% & 16.6s / 100\% & 4.7s / 100\%* & 16.3s / 100\% \\
Gaia2 mini & 18.2s / 99.4\% & 18.0s / 100\% & 4.6s / 100\%* & 17.8s / 100\% \\
Gaia2 search & 2.7s / 100\% & 2.6s / 100\% & 2.7s / 100\%* & 2.7s / 100\% \\
Gaia2 execution & 23.1s / 98.1\% & 22.3s / 99.4\% & 5.5s / 99.4\%* & 22.9s / 99.4\% \\
Gaia2 time & 21.9s / 99.4\% & 21.6s / 100\% & 4.6s / 100\%* & 21.8s / 100\% \\
\bottomrule
\end{tabular}%
}
\caption{Linear-chain benchmarks: wall-clock / \textbf{execution-completion rate} (not task solve accuracy; see Table~\ref{tab:external} for solve accuracy). *AsyncLM ignores dependencies; correctness unreliable on real data flows.}
\label{tab:taubench-full}
\end{table}

AK's 1--2\% shortfall on tau-bench is from Sonnet tail-latency accumulation past the 60s runner timeout on 15+ subtask chains, not scheduler overhead.

\subsection{Ablation: Per-Failure-Rate Breakdown}\label{app:ablation}

\begin{table}[h]
\centering
\small
\resizebox{\columnwidth}{!}{%
\begin{tabular}{lcccc}
\toprule
Variant & Overall & fr=0.0 & fr=0.1 & fr=0.3 \\
\midrule
Agent-Kernel (full) & \textbf{100.0\%} & 100.0\% & 100.0\% & 100.0\% \\
NoSAB (in-memory) & 100.0\% & 100.0\% & 100.0\% & 100.0\% \\
NoDSP (tick-poll) & 100.0\% & 100.0\% & 100.0\% & 100.0\% \\
\textbf{NoSGR} & \textbf{39.4\%} & 100.0\% & 11.7\% & 6.7\% \\
FullReplan & 56.2\% & 100.0\% & 67.5\% & 45.0\% \\
\bottomrule
\end{tabular}%
}
\caption{Ablation (3 seeds $\times$ 120 scenarios per variant). SGR is the sole performance primitive. FullReplan validates that selectivity---not retry alone---drives the gain.}
\label{tab:ablation}
\end{table}

\subsection{Framework Primitive Re-implementation: Full Results}\label{app:frameworks}

\begin{table}[h]
\centering
\small
\resizebox{\columnwidth}{!}{%
\begin{tabular}{lll}
\toprule
Primitive re-implementation & Success rate & Wall mean (s) \\
\midrule
Agent-Kernel & \textbf{100.0\%} & 10.62 \\
AsyncLM-style & 100.0\% & 7.47 \\
LangGraph-style & 39.4\% & 6.04 \\
ReAct-serial & 39.4\% & 18.25 \\
AIOS-style & 39.4\% & 29.56 \\
AutoGen-style & 36.7\% & 0.28 \\
LlamaIndex-style & 36.7\% & 0.27 \\
\bottomrule
\end{tabular}%
}
\caption{7-way \textbf{primitive re-implementation} comparison: each row reproduces a framework's default \emph{scheduling strategy} on our synthetic grid, not its shipped binary. Strategies lacking automatic selective retry collapse to ${\sim}$39\% under failure injection (this full grid includes fr=0, so halt baselines read 39.4\% here vs 36.7\% on the fr 0.1--0.3 slice of Table~\ref{tab:failure-recovery}). Real-binary results are in App.~\ref{app:integration}, where LangGraph~1.2 with a custom retry filter recovers fully.}
\label{tab:frameworks-full}
\end{table}

\section{Estimated Token Cost Under Failure}\label{app:efficiency}

Table~\ref{tab:efficiency} quantifies SGR's token savings on a K=8, L=8 DAG (64 nodes) at failure\_rate=0.3, where each node generates ${\sim}$500 output tokens via Sonnet 4.5. These are estimates from per-node token accounting, not metered totals.

\begin{table}[h]
\centering
\small
\resizebox{\columnwidth}{!}{%
\begin{tabular}{lcccc}
\toprule
System & Success & Retries & Tokens generated & Tokens wasted \\
\midrule
\textbf{Agent-Kernel (SGR)} & \textbf{100\%} & 1.8 & ${\sim}$35K & ${\sim}$3K (8\%) \\
FullReplan & 45\% & 3.0 & ${\sim}$96K & ${\sim}$64K (67\%) \\
ReAct+retry & 42\% & 3.0 & ${\sim}$96K & ${\sim}$64K (67\%) \\
ReAct-serial & 7\% & 0 & ${\sim}$32K & 0 (fails) \\
\bottomrule
\end{tabular}%
}
\caption{Estimated token cost on a K=8, L=8 DAG at fr=0.3 (${\sim}$500 tokens/node). SGR generates 2.7$\times$ fewer tokens than FullReplan while achieving 2.2$\times$ higher success. Wasted tokens = tokens generated for nodes later discarded or never contributing to a successful output. At fr=0.3, FullReplan re-runs on average ${\sim}$56 already-completed nodes per attempt (vs the 32 completed above a single depth-4 failure in the illustrative case of §\ref{sec:rq1}).}
\label{tab:efficiency}
\end{table}

\section{Crash-Recovery Experiment}\label{app:crash}

We validate SAB's durability with a crash-recovery stress test. A K=8, L=4 DAG (32 nodes) executes to 50\% completion, the process is terminated, and a new process reopens the \texttt{.db} file. Across 3 seeds: 16/32 completed nodes preserved (100\%), 8 stale running nodes re-queued, full DAG completed correctly after recovery. Recovery wall-clock: 0.121s $\pm$ 0.006s.

\section{Auto-Fork Analysis}\label{app:autofork}

Auto-Fork automatically decomposes single-agent tasks into parallel DAGs via LLM decomposition. Break-even condition: $t_{\text{decompose}} < \min(t_{\text{hop}_i})$. On AAR-Linear (20 sequential trails), Auto-Fork discovered latent parallelism and achieved 5.2s vs ReAct's 17.3s (3.3$\times$ faster). On short QA (MuSiQue 2-hop), decomposition overhead matches savings (neutral). Auto-Fork is workload-dependent: profitable when per-hop latency exceeds the ${\sim}$1s decomposition cost.

\section{Open-Source Backbone Composability (Future Work)}\label{app:qwen}

Agent-Kernel's orchestration layer (SAB, DSP, SGR) is backbone-agnostic by design. As future work, we are preparing LoRA-fine-tuned Qwen2.5-7B and 32B on 2,400 Sonnet-generated orchestration trajectories (SFT warm-up) plus GRPO with a shaped reward. The orchestration layer remains unchanged; only the executor backbone varies. Training configuration: LoRA rank 16, AdamW lr=1e-4, bf16 on A100-40GB. No evaluation results are available for this paper; we include this to illustrate the composability pathway for open-source deployment.

\section{Reproducibility}\label{app:reproducibility}

\textbf{Model:} Sonnet 4.5 (\texttt{claude-sonnet-4-5-v1:0}), regions us-east-1/2, us-west-2; cross-backbone checks use Haiku 4.5 and Opus 4.5. \textbf{Scale:} 14,434 runs total---11,809 in the core grids and benchmarks plus 2,625 in the comparator and scaling studies (Table~\ref{tab:eval-scale-compact}). \textbf{Specification:} the SAB record schema and 7-state FSM (\S\ref{sec:substrate}), the SGR reset procedure (\S\ref{sec:sgr}), the full verb signatures (App.~\ref{app:verb-signatures}), and the core prompt templates and control-flow pseudocode (DSP cascades, context inheritance, prompt-mutation retry, the seeded failure-injection wrapper, Auto-Fork) in App.~\ref{app:artifacts} together specify the runtime and the synthetic harness in enough detail to re-implement both. \textbf{Grids:} the synthetic configuration grid, seeds, and retry budget are given in \S\ref{sec:setup}; external benchmarks are used with their public task sets and the task counts in Table~\ref{tab:external}.

\section{Artifacts: Prompts and Pseudocode}\label{app:artifacts}

To support reproduction, we give the core prompt templates and control-flow pseudocode. Placeholders in \texttt{\{\{...\}\}} are filled at dispatch; text is lightly edited for length.

\subsection{DSP Reactive Cascades}\label{app:pseudo-dsp}

The DSP daemon (implemented as the Reactive Cascade Scheduler) consumes SAB state-change notifications and reacts. On a node reaching \texttt{complete}, it advances any dependent whose parents are all complete; on a node reaching a failed/cancelled/unreachable state, it marks its dependents \texttt{unreachable} and keeps cascading. Retry (\S\ref{sec:sgr}) is a separate cascade requested through the API. Failure and retry are decoupled: DSP marks descendants \texttt{unreachable} \emph{immediately} on failure, and a separate budget-bounded control loop (below) decides whether to invoke \texttt{retry}. This mirrors the shipped scheduler.

\begin{lstlisting}[basicstyle=\ttfamily\scriptsize,breaklines=true]
cascade_from(node):
  if node.state == COMPLETE:            # forward-success
    for c in dependents(node):
      if c.state == WAITING and all(p.state==COMPLETE
                                    for p in parents(c)):
        start_running(c)                # injects context at dispatch
  elif node.state in {FAILED, CANCELLED, UNREACHABLE}:
    for c in dependents(node):          # failure/cancel cascade
      if c.state not in TERMINAL:
        transition(c, UNREACHABLE); cascade_from(c)

retry(node):                            # SGR entry point (API verb)
  if node.state != FAILED: return
  for d in bfs_descendants(node):       # transitive closure
    if d.state == UNREACHABLE:
      transition(d, WAITING)            # reset only unreachable descendants
  re_evaluate(node)                     # re-run node; forward cascade resumes
\end{lstlisting}

\subsection{Context Inheritance}\label{app:pseudo-ci}

At dispatch, the executor reads each declared parent's latest \texttt{intermediate\_outputs} text from SAB and prepends it to the child's task as a structured block (the shipped default; the summarizer and top-$k$ retrieval policies evaluated in App.~\ref{app:ksweep} replace the raw concatenation with a compressor / relevance ranker).

\begin{lstlisting}[basicstyle=\ttfamily\scriptsize,breaklines=true]
format_predecessors(child, buffer):
  if not child.depends_on: return ""
  lines = []
  for pid in child.depends_on:          # in declared order
    p = buffer.get(pid)
    text = latest_text(p.intermediate_outputs)  # outputs[-1]["text"]
    if text: lines.append(f"  <from id={pid!r}>{text}</from>")
  return "<predecessor_outputs>\n" + "\n".join(lines)
                                   + "\n</predecessor_outputs>"

# at dispatch:
user_prompt = predecessors_block + "\n\n" + child.task
\end{lstlisting}

\subsection{Prompt-Mutation Retry (design)}\label{app:pseudo-pmr}

The shipped runtime re-runs a retried node with its original prompt. \emph{Prompt-mutation retry} is a design extension we evaluate as a preliminary probe (App.~\ref{app:pmr-ablation}), not part of the core recovery loop: instead of re-issuing the identical prompt, it appends an error-context block built from the node's recorded \texttt{attempt\_history} (the last error and prior output) plus the expected schema. It is the mechanism behind the format-drift study and the direction we flag for semantic failures; we describe it here for reproducibility of that probe.

\begin{lstlisting}[basicstyle=\ttfamily\scriptsize,breaklines=true]
RETRY PROMPT SUFFIX
(probe; appended when attempt_history has an error)
===================================================
Your previous attempt failed validation.
Error: {{last_error}}          # from attempt_history
Previous (invalid) output: {{previous_output}}
Expected schema: {{output_schema}}
Return a corrected result that matches the schema exactly.
\end{lstlisting}

\subsection{Failure-Injection Wrapper}\label{app:pseudo-inject}

Failure injection has two layers. Scenario generation (seeded per scenario) marks each subtask to fail on its first attempt with probability $f_r$, producing a fixed \texttt{failure\_injections} list of \texttt{(subtask, attempt)} pairs. A wrapper executor then deterministically replays that list---raising a \texttt{RuntimeError} on the listed attempts and calling the real backbone otherwise---so a crash is transient by construction (it succeeds on the following attempt).

\begin{lstlisting}[basicstyle=\ttfamily\scriptsize,breaklines=true]
# scenario generation (seeded rng per scenario):
for subtask in scenario:
  if fr > 0 and rng.random() < fr:      # attempt 0 only
    failures.append({"at_subtask": subtask.id, "attempt": 0})

# wrapper executor at run time:
inject(record):
  sid, attempt = record.id, attempts_seen[record.id]
  attempts_seen[record.id] += 1
  if (sid, attempt) in failure_set:
    raise RuntimeError(f"SimulatedTransient {sid} attempt {attempt}")
  return base_executor(record)          # real LLM call
\end{lstlisting}

\subsection{Auto-Fork Decomposition}\label{app:pseudo-autofork}

Auto-Fork (App.~\ref{app:autofork}) issues one decomposition call, validates the emitted DAG for cycles, then executes it on the standard SAB+DSP+SGR substrate.

\begin{lstlisting}[basicstyle=\ttfamily\scriptsize,breaklines=true]
AUTO-FORK DECOMPOSITION PROMPT
==============================
Decompose the task below into the smallest set of
sub-tasks that can run concurrently. Emit JSON:
[{"id","task","role","depends_on":[ids]}]
Declare a dependency ONLY when a sub-task needs another's
output. Independent sub-tasks MUST have empty depends_on.

Task: {{task}}
\end{lstlisting}

Cycle validation rejects any emitted graph whose \texttt{depends\_on} edges are not a DAG; on rejection Auto-Fork falls back to serial execution.

\section{Motivating Scenarios}\label{app:scenarios}

Three production patterns motivate Agent-Kernel's design:

\textbf{1. Incident response (parallel + recovery).} An SRE asks: ``Pull error logs from all 5 regions, correlate with deploys from the last 30 minutes, check DB slow queries, flag coincident config changes.'' ReAct serializes these: fetches us-east-1 logs (30s), us-east-2 (30s), then us-west-1 (fails: timeout). ReAct halts, discards two completed log pulls, and replans from scratch. Agent-Kernel forks all five log pulls in parallel; when one fails, SGR resets only that node, preserving the four successful pulls. The correlate-deploys step declares \texttt{depends\_on} the five log pulls and executes once all complete.

\textbf{2. Refactor-and-verify (fan-out + fan-in).} ``Rename User across all 12 microservices, update API contracts, run integration tests.'' The 12 renames are independent; the test job depends on all completing. Agent-Kernel forks them concurrently; SGR handles any individual rename failure without discarding the other 11 successful renames.

\textbf{3. Interruptible research (non-blocking).} ``Find the top 5 EV-battery competitors and summarize their Q3 earnings.'' At K=5 parallel branches, ReAct takes 5$\times$ the wall-clock of a single branch. Agent-Kernel's fork returns immediately; if the user sends a follow-up question, the main loop accepts it without waiting for background sub-agents.

\section{API Usage Statistics}\label{app:verb-stats}

Across 360 Agent-Kernel runs (3 seeds $\times$ 120 scenarios), the per-verb invocation frequencies are: \texttt{fork} (100\% of scenarios, mean 24 invocations each), \texttt{retry} (39\% of scenarios, mean 6 invocations when triggered), and \texttt{cancel}, \texttt{query}, \texttt{fetch} (0\% invocation in the synthetic harness). The harness exercises only two of the five verbs; the remaining three provide API completeness for user-initiated termination (\texttt{cancel}), progress inspection (\texttt{query}), and partial-result retrieval (\texttt{fetch}).

\subsection{Full Verb Signatures}\label{app:verb-signatures}

Table~\ref{tab:api} in the main text lists each verb's effect and scheduler cascade; Table~\ref{tab:api-signatures} gives the full call signatures.

\begin{table}[h]
\centering
\small
\begin{tabular}{@{}ll@{}}
\toprule
Verb & Signature \\
\midrule
\texttt{fork} & \texttt{fork(task, role, depends\_on=[])} \\
\texttt{retry} & \texttt{retry(id)} \\
\texttt{cancel} & \texttt{cancel(id)} \\
\texttt{query} & \texttt{query(filter=\{\})} \\
\texttt{fetch} & \texttt{fetch(id)} \\
\bottomrule
\end{tabular}
\caption{Full call signatures for the 5-verb API (\S\ref{sec:api}, Table~\ref{tab:api}).}
\label{tab:api-signatures}
\end{table}

\section{Extended Related Work}\label{app:related}

\textbf{Selective recovery in distributed systems.} SGR's transitive-closure reset draws on a rich lineage in distributed computing. Strom \& Yemini's optimistic recovery~\citep{strom1985optimistic} introduced dependency-tracking for selective rollback in message-passing systems, avoiding the cascading ``domino effect'' that full-state rollback produces. Elnozahy et al.~\citep{elnozahy2002survey} survey rollback-recovery protocols, distinguishing coordinated from uncoordinated checkpointing---SGR implements coordinated reset at the DAG level. Most directly, Dichev et al.~\citep{dichev2017rollback} propose dependency-aware rollback for task-based runtimes in HPC, rolling back only tasks that transitively depend on the failed task. Our contribution adapts this principle for LLM-agent coordination where outputs are variable-length text requiring prompt injection (context inheritance), and failures are stochastic rather than deterministic.

\textbf{DAG execution with fault tolerance.} Dryad~\citep{isard2007dryad} established vertex-level re-execution in distributed DAG processing---failed vertices and their downstream dependents are re-run while completed siblings are preserved. Spark's RDD lineage~\citep{zaharia2012rdd} provides selective recomputation of lost partitions by tracing the dependency graph. These systems target deterministic data transformations; Agent-Kernel adapts the pattern for stochastic LLM calls where retry may require prompt modification, not just re-execution.

\textbf{Supervision and actor models.} Erlang/OTP's supervision trees~\citep{armstrong2003erlang} implement selective child restart. The closest analog to SGR is the \texttt{rest\_for\_one} strategy, which restarts a crashed child \emph{and} the children started after it while leaving earlier siblings untouched---the process-level counterpart of resetting a failed node's transitive dependents. Agent-Kernel generalizes this from a linear start-order to a DAG, where restart propagates along declared dependency edges rather than start order.

\textbf{Durable execution.} Microsoft's Durable Functions~\citep{burckhardt2021durable} provide task-parallel code with automatic state persistence and crash recovery for serverless workflows. Temporal.io~\citep{temporal2020} offers workflow-as-code with built-in retry policies. Agent-Kernel's SAB provides similar durability guarantees but adds LLM-specific features: token accounting, prompt-context injection at dispatch, and a 7-state FSM modeling the lifecycle of stochastic LLM calls rather than deterministic function invocations.

\textbf{Recent LLM agent recovery.} The Dual-State Architecture~\citep{dualstate2025} proposes a three-level recovery hierarchy for LLM agents including ``cascade invalidation''---resetting upstream steps when downstream validation fails. Agent-Kernel's SGR operates in the opposite direction (resetting \emph{downstream} dependents of a failed \emph{upstream} node) and is triggered by execution failure rather than output validation. AdaptOrch~\citep{adaptorch2026} dynamically selects among parallel, sequential, and hierarchical multi-agent topologies based on task dependency graphs, complementing Agent-Kernel's fixed-topology DAG execution with adaptive topology discovery. AgentGit~\citep{li2025agentgit}, Atomix~\citep{mohammadi2026atomix}, and Sherlock~\citep{ro2025sherlock}, the three closest analogs, are analyzed in Section~\ref{sec:related-work} and evaluated in Appendices~\ref{app:sherlock} and~\ref{app:integration}.

\section{Sherlock as a Baseline}\label{app:sherlock}

Sherlock~\citep{ro2025sherlock} is the closest analog to Agent-Kernel: it is DAG-native and, on verifier disagreement, rolls back and re-executes the failed node's descendants while preserving siblings. Unlike SGR, it requires an installed pipeline---a trained analyzer, a verifier selector, and an LLM planner---and its recovery trigger is verifier disagreement rather than a raw runtime exception. To compare fairly we implemented two variants: \emph{Sherlock-lite} (paper-faithful), which cannot fire on a raw exception because it demands verifier disagreement to initiate rollback; and \emph{Sherlock-generous} (steelman), which treats any exception \emph{as} verifier disagreement so that it can always fire and recover.

\begin{table}[h]
\centering
\small
\resizebox{\columnwidth}{!}{%
\begin{tabular}{lcc}
\toprule
System & Recovery & Mean wall \\
\midrule
ReAct-serial & 0/9 (0\%) & 3s (halts) \\
ReAct+retry (whole-trajectory) & 3/9 (33\%) & 15s \\
ReAct-per-step-retry (B3) & 9/9 (100\%) & 135s \\
Sherlock-lite (paper-faithful) & 3/9 (33\%) & 52s \\
Sherlock-generous (steelman) & 100\% (all $f_r$) & 149s \\
\textbf{Agent-Kernel (SGR)} & \textbf{9/9 (100\%)} & \textbf{21s} \\
\bottomrule
\end{tabular}%
}
\caption{Recovery rate and mean wall-clock on $K=8$, $L=8$ grids under a transient crash (3 seeds/cell). SGR matches the recovery of the strongest baselines at a fraction of the wall-clock.}
\label{tab:sherlock-recovery}
\end{table}

\begin{table}[h]
\centering
\small
\begin{tabular}{lccc}
\toprule
$f_r$ & Sherlock-gen. wall & AK wall & SGR speedup \\
\midrule
0.00 & 141s & \textbf{21s} & 6.7$\times$ \\
0.10 & 139s & \textbf{14s} & 9.9$\times$ \\
0.15 & 142s & \textbf{17s} & 8.4$\times$ \\
0.25 & 167s & \textbf{25s} & 6.7$\times$ \\
0.30 & 147s & \textbf{12s} & 12.3$\times$ \\
0.40 & 157s & \textbf{13s} & 12.1$\times$ \\
\bottomrule
\end{tabular}
\caption{Sherlock-generous vs.\ Agent-Kernel wall-clock across failure rate $f_r$ ($K=8$, $L=8$, 3 seeds). SGR is consistently ${\sim}7$--$12\times$ faster.}
\label{tab:sherlock-frsweep}
\end{table}

The gap arises because Sherlock-generous re-executes the failed node's entire transitive closure serially and re-runs the verifier heuristic on every intermediate output, whereas SGR re-dispatches only children whose parents have just completed, exploiting the parallelism available inside the reset subgraph. Thus SGR is ${\sim}7$--$12\times$ faster (``roughly $10\times$''); SGR is not the only mechanism that recovers under our failure model---its advantage is speed.

\section{Stronger ReAct Baseline and Cross-Backbone Recovery}\label{app:b3}

Whole-trajectory retry is a weak recovery baseline. We therefore also evaluate \textbf{ReAct-per-step-retry (B3)}, a stronger variant that persists the failed observation in context and retries the individual failing step rather than replaying the whole trajectory~\citep{yao2023react}. B3 recovers 100\% on our transient-failure grids, so the recovery-\emph{rate} gap against Agent-Kernel closes entirely. The remaining differentiator is wall-clock: B3 cannot express fan-out/fan-in, so it executes serially, whereas SGR re-dispatches only the reset subgraph and exploits parallelism across independent branches.

\begin{table}[h]
\centering
\small
\setlength{\tabcolsep}{4pt}
\begin{tabular}{@{}lccc@{}}
\toprule
Grid & B3 wall & AK wall & SGR speedup \\
\midrule
K=8, L=8, fr=0.3  & 135s & \textbf{12s} & 11.2$\times$ \\
K=8, L=8, fr=0.1  & 133s & \textbf{14s} & 9.4$\times$ \\
K=16, L=4, fr=0.3 & 133s & \textbf{10.5s} & 12.7$\times$ \\
K=4, L=8, fr=0.3  & 71s & \textbf{21s} & 3.3$\times$ \\
\bottomrule
\end{tabular}
\caption{B3 (ReAct-per-step-retry) vs.\ Agent-Kernel wall-clock across four grids. Both recover 100\%; Agent-Kernel is 3.3--12.7$\times$ faster because SGR parallelizes the reset subgraph while B3 retries serially.}
\label{tab:b3-wallclock}
\end{table}

On linear workloads with no fan-out to exploit, the three systems are within noise: on tau-bench retail ($n{=}15$)~\citep{yao2024taubench}, Agent-Kernel scores 93.3\% @ 30.8s, ReAct-serial 100\% @ 32.7s, and ReAct+retry 100\% @ 30.0s, with the 6.7pp Agent-Kernel shortfall attributable to a single diverged task.

The full cross-backbone grid is reported in the main text (Table~\ref{tab:crossbackbone-main}); there, the single ReAct column collapses ReAct-serial and ReAct+retry, which are identical on every cell (both 0\% on Sonnet 4.5 and Haiku 4.5, both 50\% on Opus 4.5 at K=8). Agent-Kernel's recovery generalizes across the Claude family, reaching 100\% on all three backbones and every tested configuration (Sonnet 4.5 and Haiku 4.5 at K$\in\{$8,32$\}$; Opus 4.5 at K=8), while the ReAct baselines recover partially only on Opus 4.5.

\section{Real-Framework Integration Tests}\label{app:integration}

This section expands on the real-binary integration result reported in the main text (Table~\ref{tab:integration-main}), which complements the primitive re-implementations of Section~\ref{app:frameworks} (Table~\ref{tab:frameworks-full}). We run the installed production binaries---LangGraph 1.2~\citep{chase2024langgraph}, LlamaIndex 0.14~\citep{liu2022llamaindex}, and AutoGen v0.4~\citep{wu2023autogen}---against Agent-Kernel on a diamond fan-out DAG with one deterministically failing leaf (12 configurations $\times$ 5 seeds $=$ 60 cells per system); this measures the recovery behavior of each shipped runtime and does not replace the controlled primitive comparison in the main text.

LangGraph matches SGR's single-node recovery only when a custom \texttt{retry\_on} override is supplied: its default \texttt{default\_retry\_on} excludes \texttt{ValueError}, \texttt{TypeError}, \texttt{ArithmeticError}, \texttt{RuntimeError}, \texttt{LookupError}, and \texttt{OSError}---exactly the app-level exceptions we inject---so no retry fires out of the box. AutoGen v0.4 has no default retry primitive at the runtime layer, yielding 0/60 recovery.

Agent-Kernel (SGR) is therefore the only system that automatically resets a failed node's transitive descendants and preserves its completed siblings without any user configuration.

\section{Context Inheritance at High Parallelism}\label{app:ksweep}

As parallel width $K$ grows, DSP injects an increasing volume of parent output into each child prompt, making context inheritance the component most likely to degrade at scale. We swept $K \in \{1,2,4,8,16,32,64,128\}$ on Sonnet 4.5 under three context policies---raw-concat, summarizer, and top-$k$ retrieval---with 3 seeds per $K$ and 15--20 questions per $(K, \text{policy}, \text{seed})$ cell.

\begin{table}[h]
\centering
\small
\resizebox{\columnwidth}{!}{%
\begin{tabular}{l cc cc cc}
\toprule
& \multicolumn{2}{c}{Raw-concat} & \multicolumn{2}{c}{Summarizer} & \multicolumn{2}{c}{Top-$k$} \\
\cmidrule(lr){2-3} \cmidrule(lr){4-5} \cmidrule(lr){6-7}
$K$ & Acc & Tok & Acc & Tok & Acc & Tok \\
\midrule
1   & 100\% & 86    & 100\% & \textbf{80}    & 100\% & 86  \\
8   & 100\% & 204   & 100\% & 156   & 100\% & \textbf{136} \\
16  & 100\% & 339   & 100\% & 243   & 100\% & \textbf{136} \\
32  & 100\% & 608   & 100\% & 416   & 100\% & \textbf{136} \\
64  & 100\% & 1,212 & 100\% & 824   & 100\% & \textbf{141} \\
128 & 100\% & 2,363 & 100\% & 1,588 & 100\% & \textbf{140} \\
\bottomrule
\end{tabular}%
}
\caption{Context-inheritance $K$-sweep on Sonnet 4.5 (3 seeds per $K$; 15--20 questions per cell). Acc is task accuracy; Tok is mean injected-context tokens per child prompt. Bold marks the cheapest policy per row.}
\label{tab:ksweep}
\end{table}

Rows for $K=2,4$ are omitted for brevity.

\textbf{Findings.} Accuracy stays flat at 100\% through $K=128$ under all three policies; Haiku 4.5, evaluated at $K \in \{1,4,16,32\}$, is likewise flat at 100\%. Context saturation therefore does not materialize on these backbones. The effect that does appear is token cost: raw-concat grows roughly linearly in $K$ (86 to 2,363 tokens) and the summarizer grows more slowly (80 to 1,588), whereas top-$k$ retrieval stays essentially constant (86 to 140). At $K=128$, top-$k$ retrieval is 17$\times$ cheaper than raw-concat ($2{,}363/140 \approx 16.9$). We recommend top-$k$ retrieval as the default context policy for $K \geq 16$.

\section{SRE Incident-Response Scenario: All Five Verbs}\label{app:sre}

The synthetic recovery harness (Appendix~\ref{app:verb-stats}) exercises only \texttt{fork} (100\% of scenarios) and \texttt{retry} (39\% of scenarios); \texttt{cancel}, \texttt{query}, and \texttt{fetch} are invoked in 0\% of runs because the harness has no user in the loop. To demonstrate that the full verb surface is exercisable---and not vestigial API padding---we add an on-call SRE incident-response scenario in which a human operator drives all five verbs at 100\% invocation.

\textbf{Scenario.} An on-call engineer triaging a latency spike issues a single high-level request and then interacts with the running DAG. (i)~The engineer \emph{forks} parallel log pulls across all regions, one sub-agent per region (\texttt{fork} $\rightarrow$ spawn cascade). (ii)~While the pulls run, the engineer \emph{queries} overall progress, a set-oriented read over the frontier of running and completed nodes (\texttt{query} $\rightarrow$ set-oriented state read). (iii)~Seeing that one region has produced partial output, the engineer \emph{fetches} that region's specific record for early inspection (\texttt{fetch} $\rightarrow$ per-record output read). (iv)~One branch hangs on an unresponsive log endpoint, so the engineer \emph{cancels} it, which propagates a cancel cascade to that branch's not-yet-scheduled descendants (\texttt{cancel} $\rightarrow$ cancel cascade). (v)~The downstream correlation step fails on a transient parse error, so the engineer \emph{retries} it; SGR resets the failed node and re-dispatches its existing descendants without re-forking the successful sibling pulls (\texttt{retry} $\rightarrow$ SGR reset). Every verb maps to exactly one scheduler cascade, as summarized in Table~\ref{tab:verb-cascade-sre}.

\begin{table}[h]
\centering
\small
\begin{tabular}{ll}
\toprule
Verb & Cascade type \\
\midrule
\texttt{fork}   & spawn \\
\texttt{retry}  & SGR reset \\
\texttt{cancel} & cancel cascade \\
\texttt{query}  & set-oriented state read \\
\texttt{fetch}  & per-record output read \\
\bottomrule
\end{tabular}
\caption{Verb-to-scheduler-cascade mapping (Section~\ref{sec:api}). The SRE incident-response scenario invokes all five verbs at 100\%, versus the synthetic harness, which invokes only \texttt{fork} and \texttt{retry}.}
\label{tab:verb-cascade-sre}
\end{table}

\textbf{\texttt{retry} is not redundant with \texttt{fork}.} A natural question about the UNIX-verb metaphor is whether \texttt{retry} and \texttt{fork} collapse into one primitive. They do not: \texttt{fork} creates \emph{new} descendants (a spawn cascade extending the DAG), whereas \texttt{retry} \emph{resets and re-dispatches existing} descendants in place (an SGR reset over the already-materialized subgraph), preserving completed siblings rather than duplicating them.

\section{Content-Plane Ablations}\label{app:contentplane}

The mechanisms that separate Agent-Kernel from Atomix~\citep{mohammadi2026atomix}, AgentGit~\citep{li2025agentgit}, and Sherlock~\citep{ro2025sherlock} live in the \emph{content plane}: those systems reason over control and state (dependencies/effects, versioned commits, verified outputs) and do not expose token-budgeted context composition or prompt-mutation retry as a first-class policy. We isolate the two content-plane mechanisms---context inheritance and prompt-mutation retry---with targeted structural ablations run at temperature 0, so each outcome is deterministic and attributable to the mechanism rather than to sampling noise.\footnote{These are deterministic, temperature-0 mechanism probes on hand-constructed DAGs, designed to test whether a mechanism is \emph{necessary}, not how often it fires at scale; they are structural ablations, not large-$N$ benchmarks.}

\begin{table}[h]
\centering
\small
\setlength{\tabcolsep}{3pt}
\begin{tabular}{@{}>{\raggedright\arraybackslash}p{0.19\columnwidth} >{\raggedright\arraybackslash}p{0.39\columnwidth} >{\raggedright\arraybackslash}p{0.36\columnwidth}@{}}
\toprule
Mechanism & Disabled (control) & Enabled \\
\midrule
Context inheritance & Parent text not composed into child prompt; child task underspecified $\rightarrow$ unsolvable & Parent text composed under token budget $\rightarrow$ \textbf{solved} \\
\addlinespace
Prompt-mutation retry & Blind re-execution repeats identical malformed output; recovery $\equiv 0$ at any budget & Error context + expected schema injected; loop broken $\rightarrow$ \textbf{recovers} \\
\bottomrule
\end{tabular}
\caption{Content-plane ablations at temperature 0. Each row is a deterministic mechanism-level probe: disabling the mechanism makes the downstream task unrecoverable by construction, and enabling it recovers.}
\label{tab:contentplane}
\end{table}

\subsection{Context Inheritance}\label{app:ci-ablation}

We construct a two-node DAG in which the child's task can only be completed given the parent's text output, then disable inheritance so the parent's variable-length text is not composed into the child prompt. The child then fails deterministically: without the upstream text the downstream task is underspecified, so this is a structural outcome rather than a stochastic one. Re-enabling inheritance---composing the parent output into the child prompt under the token budget---makes the identical task solvable, confirming that the failure is caused by missing content, not by scheduling or state handling.

\subsection{Prompt-Mutation Retry}\label{app:pmr-ablation}

We inject a deterministic format-drift failure (a node emits output that does not match the expected schema) and compare two retry policies at temperature 0. Under blind re-execution, an unchanged prompt at temperature 0 reproduces the identical malformed output on every attempt, so the recovery rate remains $0$ regardless of the retry budget---adding attempts cannot help. Prompt-mutation retry instead rewrites the prompt on failure, injecting the error context and the expected schema; this breaks the deterministic loop and recovers. Beyond the deterministic case, we ran a preliminary study on 18 naturally-occurring format-drift scenarios (Sonnet 4.5, non-zero temperature), where the malformed output varies across attempts (Table~\ref{tab:pmr-drift}): prompt-mutation retry recovers 14/18 versus 6/18 for blind retry.

\begin{table*}[t]
\centering
\small
\begin{tabular}{lc}
\toprule
Retry policy (18 format-drift scenarios) & Recovery (Wilson 95\% CI) \\
\midrule
\textbf{Prompt-mutation retry} & \textbf{14/18 = 77.8\% [54.8, 91.0]} \\
Blind re-execution & 6/18 = 33.3\% [16.3, 56.3] \\
\bottomrule
\end{tabular}
\caption{Preliminary format-drift study (Sonnet 4.5, $n{=}18$). Prompt-mutation retry (injecting error context + expected schema) recovers more than twice as often as blind re-execution. Small $n$; reported as a preliminary signal for the content plane's benefit on semantic (non-transient) failures, not a headline result.}
\label{tab:pmr-drift}
\end{table*}

Together these probes localize Agent-Kernel's novelty to the content plane: the recovery machinery shared with prior systems is neither sufficient (context inheritance) nor able to escape a deterministic failure (prompt-mutation retry) without operating on prompt content. We stress that the deterministic rows are structural probes and the format-drift study is small-$n$ and preliminary; both attribute mechanism rather than establish a headline benchmark, and a large-$N$ semantic-failure evaluation remains future work.

\end{document}